\documentclass{article}

\newcommand\KL[2]{{\mathrm{KL}}\left(#1\,||\,#2\right)}
\newcommand\expectation[2]{\mathsf{E}_{#1}\left[#2\right]}

\newcommand{\trans}{W}

\usepackage{bbm}
\usepackage[T1]{fontenc}    % use 8-bit T1 fonts
\usepackage{url}            % simple URL typesetting
\usepackage{booktabs}       % professional-quality tables
\usepackage{amsfonts,amsmath,amssymb,stmaryrd}
\usepackage{nicefrac}       % compact symbols for 1/2, etc.
\usepackage[capitalise]{cleveref}   
\usepackage{microtype}      % microtypography
\usepackage{algorithm} 
\usepackage{algorithmic} 
\usepackage{nicefrac}       % compact symbols for 1/2, etc.
\usepackage{mathtools}
\usepackage{tabu}
\usepackage{amsthm}
\usepackage{wrapfig}
\usepackage{ bbold }
\usepackage{mathabx}

\usepackage[shortlabels,inline]{enumitem}  %inline enumeration
\newlist{inlineitemize}{enumerate*}{1}
\setlist[inlineitemize]{label=(\roman*)}

\theoremstyle{remark}

\usepackage[accepted]{icml2021}

\icmltitlerunning{Active Learning of Continuous-time Bayesian Networks through Interventions}

\begin{document}

\twocolumn[
\icmltitle{Active Learning of Continuous-time Bayesian Networks through \\Interventions}




\begin{icmlauthorlist}
\icmlauthor{Dominik Linzner}{etit,kausa}
\icmlauthor{Heinz Koeppl}{etit,bio}
\end{icmlauthorlist}

\icmlaffiliation{etit}{Department of Engineering and Information Technology, TU Darmstadt, Germany}
\icmlaffiliation{bio}{Department of Biology, TU Darmstadt, Germany}
\icmlaffiliation{kausa}{The Why Company GmbH, Berlin, Germany}

\icmlcorrespondingauthor{Dominik Linzner}{dlinzner90@gmail.com}
\icmlcorrespondingauthor{Heinz Koeppl}{heinz.koeppl@bcs.tu-darmstadt.de}

\icmlkeywords{Active Learning, Structure Learning, Time-series}

\vskip 0.3in
]



\printAffiliationsAndNotice{\icmlEqualContribution} % otherwise use the standard text.

\begin{abstract}
  We consider the problem of learning structures and parameters of Continuous-time Bayesian Networks (CTBNs) from time-course data under minimal experimental resources. In practice, the cost of generating experimental data poses a bottleneck, especially in the natural and social sciences. A popular approach to overcome this is Bayesian optimal experimental design (BOED). However, BOED becomes infeasible in high-dimensional settings, as it involves integration over all possible experimental outcomes. We propose a novel criterion for experimental design based on a variational approximation of the expected information gain. We show that for CTBNs, a semi-analytical expression for this criterion can be calculated for structure and parameter learning. By doing so, we can replace sampling over experimental outcomes by solving the CTBNs master-equation, for which scalable approximations exist. This alleviates the computational burden of integrating over possible experimental outcomes in high-dimensions. We employ this framework in order to recommend interventional sequences. In this context, we extend the CTBN model to conditional CTBNs in order to incorporate interventions. We demonstrate the performance of our criterion on synthetic and real-world data.
\end{abstract}

Learning directed dependencies in multivariate data is a fundamental problem in science and has application across many disciplines such as in the natural and social sciences, finance, and engineering~\cite{Acerbi2014,Schadt2005}. However, large amounts of data are needed in order to learn these dependencies. This is a problem when data is acquired under limited resources, which is the case in dedicated experiments, e.g. in molecular biology or psychology~\cite{Steinke2007,6426738,Liepe2013,JayI.MyungandMarkA.Pitt2015,Dehghannasiri2015,10.1145/3233188.3233217}. Active learning schemes pave a principled way to design sequential experiments such that the required resources are minimized.

The framework of Bayesian optimal experimental design (BOED) \cite{ChalonerK.Verdinelli1987,Ryan2016} allows for the design of active learning schemes, which are provably~\cite{Lindley1956,Sebastiani2000} one-step optimal. However, BOED becomes infeasible in high-dimensional settings, as it involves integration over possible experimental outcomes. 

While active learning schemes have been previously applied in order to learn dependency structures~\cite{Tong2001,Eaton2007,He2008,Lindgren2018} or parameters~\cite{Rubenstein2017} of probabilistic graphical models from snapshot or static data, active learning schemes for longitudinal and especially temporal data is as of yet under-explored. 
Dynamic Bayesian networks offer an appealing framework to formulate structure learning for temporal data within the graphical model framework~\cite{Koller2010}. The fact that the time granularity of the data can often be very different from the actual granularity of the underlying process motivates the extension to continuous-time Bayesian networks (CTBN)~\cite{Nodelman1995}, where no time granularity of the unknown process has to be assumed.

In this manuscript, we present an active learning scheme for learning CTBNs from interventions. We make the following contributions: \begin{inlineitemize}\item We derive a criterion for active learning suited for the case when the space of possible experimental outcomes is high-dimensional in Section 2. \item In Section 3, we extend CTBNs to incorporate interventions, thereby introducing conditional CTBNs (cCTBNs). \item We discuss pooling of interventional data in cCTBNs in Section 4. \item We derive semi-analytical expressions of our design criterion from Section 2 for parameter and structure learning in Section 5. \item We demonstrate the performance of our approach on synthetic and real-world data in section 6 \end{inlineitemize}.

\section{Background}
 \textbf{Interventions.} Consider a directed acyclic graph  $\mathcal{G}=(\mathcal{V},\mathcal{E})$ with nodes $\mathcal{V}\equiv\{1,\dots,N\}$ and edges $\mathcal{E}\subseteq \mathcal{V}\times \mathcal{V}$. The parent-set of node $n$ is then defined as $\mathrm{par}^\mathcal{G}(n)\equiv\{m \mid  (m,n)\in \mathcal{E}\}$. Conversely, we define the \emph{child-set} $\mathrm{ch}^\mathcal{G}(n)\equiv\{m \mid  (n,m)\in \mathcal{E}\}$. Consider a joint distribution over a set of random variables over a countable domain $p(X_1,\dots,X_N)$, characterized by a Bayesian network with graph $\mathcal{G}$, i.e., 
 \begin{align*}
 p(X_1,\dots,X_N) = \prod_{i=1}^N p(X_i\mid X_{\mathrm{par}^\mathcal{G}(i)}).
 \end{align*}
Interventions as popularized in \cite{Pearl2000} are denoted via a $\mathrm{do}$-operation and correspond to a change to the model. In our example, an intervention on variable $X_k$, $\mathrm{do}(X_k=x_k)$ would change the distribution to 
\begin{align}\label{eq:do}
 p(X_1,\dots,X_N\mid \mathrm{do}(X_k=x_k) ) = \mathbb{1}(X_k=x_k)\tilde{p}(X_{\neg k} ),% \nonumber 
\end{align}
with $\mathbb{1}(\cdot)$ the indicator function and $\tilde{p}(X_{\neg k})\equiv \prod_{i=1,i\neq k}^N p(X_i\mid X_{\mathrm{par}^\mathcal{G}(i)})$, where notation $\neg k$ means all variables except $X_k$. It corresponds to a model where the conditional distribution of every variable remains unchanged, but the value of variable $X_k=x_k$ is fixed \cite{Pearl2000,Spirtes2010}. This $\mathrm{do}$-operation contrasts traditional conditioning
\begin{align*}
 p(X_{\neg k}\mid X_k=x_k ) = \frac{p(X_k=x_k\mid X_{\mathrm{par}^\mathcal{G}(k)})}{p(X_k=x_k)}\tilde{p}(X_{\neg k}),
\end{align*}
 %with $X_{-k}$ being all variables without $X_k$, 
 where conditioning on $X_k=x_k$ also affects the parents of node $k$ (instead of only its children).
 Interventions can also be modelled as external condition variables~\cite{Pearl2000} whose effects are equivalent to $\mathrm{do}$-operations. A problem that is encountered when learning from interventions is data-pooling~\cite{Eberhardt2006}: Under what conditions can observations gathered under different interventions be used to learn about the original model. We later show that CTBNs allow for data-pooling naturally.
 
% Lastly, we mention the existence of \emph{soft} interventions, which result in
  %\begin{align}
 %p(X_1,\dots,X_N\mid \mathrm{do}(X_k) ) = \tilde{p}(X_k\mid X_{\mathrm{par}(k)})\prod_{i=1,i\neq k}^N p(X_i\mid X_{\mathrm{par}(i)}).
 %\end{align}
\textbf{Conditional Continuous-time Markov Chain.}
We introduce a conditional Continuous-time Markov process (cCTMC) \cite{Nodelman1995,norris_1997} by a tuple $(\mathcal{S},\mathcal{I},\trans,s_0)$. It defines a Markov process $\{S(t)\}_{t\geq0}$ through a transition intensity matrix ${{\trans}:\mathcal{S}\times\mathcal{S}\times\mathcal{I}\rightarrow\mathbb{R}}$ over a countable state space $\mathcal{S}$ given a countable space of external conditions (interventions) $\mathcal{I}$ and initial states $s_0$, which may also be dependent on the external condition ${s_0:\mathcal{I}\rightarrow\mathcal{S}}$.  For the sake of conciseness, we will often adopt shorthand notations of the type ${p_{t'-t}(s'\mid s,i)\equiv p(S(t')=s'\mid S(t)=s,i)}$, with ${s,s'\in\mathcal{S},\,i\in\mathcal{I}}$.
 Given a condition, its time evolution can be understood as a usual continuous-time Markov chain (CTMC) with the (infinitesimal) transition probability
${
p_{{h}}(s'\mid s,i)=\mathbb{1}(s=s')+{h}\: {{\trans}}(s, s', i)+o({h}),
}$
 for some time-step ${h}$ with  ${\lim_{h\rightarrow 0}o(h)/h=0}$. We note that any intensity matrix ${{\trans}}$ fulfills ${{{\trans}}(s, s,i)=-\sum_{s'\neq s}{{\trans}}(s, s',i)}$ for any condition $i$. %To avoid clutter, we will frequently omit the dependency on $\pi$ in the marginal transition model. 
 In the continuous-time limit $h\rightarrow0$, the cCTMCs marginal probabilities can be shown to follow the Chapman--Kolmogorov-, or master-equation
\begin{align}\label{eq:m-eq}
 \frac{\mathrm d}{\mathrm d t}p_t(s\mid s_0,i) &= \sum_{s'\neq s}\left[\trans(s',s,i)p_t(s'\mid s_0,i)\right.\\
 &\nonumber\left.-\trans(s,s',i)p_t(s\mid s_0,i)\right].
\end{align}
\textbf{Bayesian Optimal Experimental Design.}
The objective of BOED is to find the design (intervention $i$) that maximizes the expected information gain $\mathrm{EIG}(i)$ about different models $\Theta$ \cite{Lindley1956,ChalonerK.Verdinelli1987}. In our case $\Theta$ will be rate matrices $W$, or their induced graph-structures, of continuous-time Markov processes (see also Section~\ref{sec:CTBN}).  The $\mathrm{EIG}(i)$ corresponds to the expected Kullback--Leibler (KL) divergence between prior $p(\Theta)$ and posterior $p(\Theta\mid D)$ after intervention $i$ under uncertain experimental outcome (data) $D$. The objective of this task can then be formulated as
\begin{align}
 i^*&=\arg\max_{i\in \mathcal{I}} \mathrm{EIG}(i),\\
\nonumber\mathrm{EIG}(i)&=\expectation{}{\KL{p(\Theta\mid D,i)}{p(\Theta)}} \label{eq:EIG},
\end{align}
with the expectation taken with respect to $p(D\mid i)$.
%This forms a design strategy that is (one-step) optimal from an information-theoretic viewpoint \cite{Lindley1972,Sebastiani2000}. 
%This design admits an information-theoretic interpretation where the experiment defines a $i$-parametrized channel $\Theta \rightarrow D$ with channel matrix $p(D \mid \Theta, i)$. Rearranging \eqref{eq:EIG} yields that $\mathrm{EIG}(i)$ coincides with the mutual information (MI) $\mathrm{I}(D,\Theta \mid i)$ and that, accordingly, the optimal design $i^*$ achieves the Shannon capacity of this channel for a given prior $p(\Theta)$. Learning then corresponds to ``decoding'' $\Theta$ from data $D$ and is known to yield minimal error if the channel was tuned to its capacity $\mathrm{EIG}(i^*)$.        
Unfortunately, in practice, \eqref{eq:EIG} is notoriously hard to evaluate \cite{Foster2019} rendering it impractical for our purposes. 
\begin{figure*}[t]
\begin{centering}
        \includegraphics[width=1.9\columnwidth]{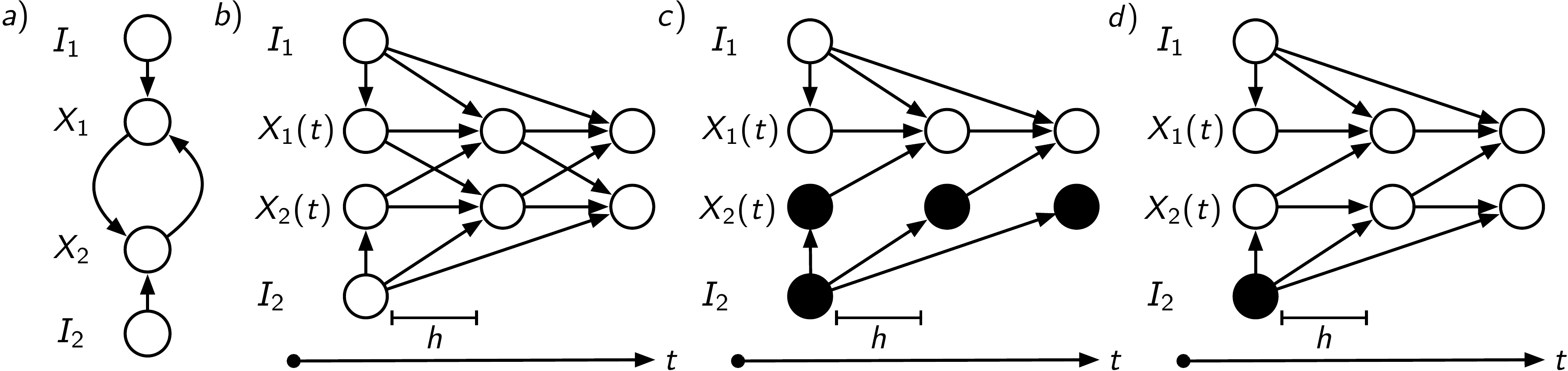}
    \caption{a) Conditional CTBN with two nodes $X_1$ and $X_2$ and local interventions $I_1$ and $I_2$. Black means for interventions $I_n\neq 0$ and for $X_n(t)=x_n^0$ for all $t$. b), c) and d) are the $h$-discretized (unrolled) cCTBN for different interventions. b) Unrolled cCTBN without intervention. c) Unrolled cCTBN with perfect intervention on $X_2$ and d) with imperfect intervention.}
    \label{fig-conditional-ctbn}
    \end{centering}
\end{figure*}

\section{Variational Box-Hill Criterion for Active Learning}\label{section-VBHC}
Evaluating the EIG is intractable, due to the repeated model posterior evaluations for all possible outcomes $D$. For this reason, various approximation techniques have been employed \cite{Lewi2009,Rainforth2018,Foster2019}. Recently, promising advances via variational approximations of the EIG have been made~\cite{Foster2019}. In the following, we want to take a similar route, while making use of our model assumptions. One way to do this is to upper-bound the EIG, which can be rewritten as
${
\mathrm{EIG}(i)=\expectation{}{\ln\frac{p(D\mid\Theta,i)}{p(D\mid i)}},
}$
with the expectation with respect to $p(D,\Theta\mid i)$.
By lower-bounding the marginal likelihood ${\ln p(D\mid i)\geq\expectation{}{\ln p(D\mid\Theta,i)}}-\mathrm{{KL}}\left[q_{\kappa}(\Theta)||p(\Theta)\right]$, the expectation subject to some, yet unspecified, variational distribution $q_{\kappa}(\Theta)$ over models, one arrives at an upper-bound to the EIG, which takes the form of a weighted KL-divergence between different possible models $\Theta$
\begin{align}
\mathrm{{EIG}}(i)&\leq\expectation{}{\expectation{}{\KL{p(D\mid\Theta,i)}{p(D\mid\Theta',i)}}}\\
&\nonumber+\KL{q_{\kappa}(\Theta)}{p(\Theta)}\equiv \mathrm{VBHC}(i,\kappa).
\end{align}
Here the outer expectation is w.r.t $p(\Theta)$, the inner one w.r.t $q_{\kappa}(\Theta')$.
We will refer to this quantity as the \emph{Variational Box-Hill Criterion} (VBHC). By setting $q_\kappa(\Theta)= p(\Theta)$, which we emphasize is \emph{not} the minimizer of the VBHC (and thus not the best approximation to the EIG), one recovers the classical Box-Hill criterion (BHC) \cite{Box1967}, ${\mathrm{BHC}(i)\equiv\expectation{}{\expectation{}{\KL{p(D\mid\Theta,i)}{p(D\mid\Theta',i)}}}}$, with expectations w.r.t $p(\Theta)$ and $p(\Theta')$. The BHC has been used in different contexts~\cite{Reilly1970,Daniel1996,JayI.MyungandMarkA.Pitt2015,Ng2004} as a criterion for design for model discrimination experiments, as it can be computed analytically for some models (e.g. for Gaussian distributions). To our surprise, it hasn't received much attention otherwise. In contrast to the BHC, the VBHC allows for minimization, and a thereby tightening the upper-bound w.r.t $\kappa$ before selecting an intervention
\begin{align}\label{eq:int-VBHC}
    i^*=\arg\max_{i\in \mathcal{I}} \min_\kappa\mathrm{VBHC}(i,\kappa).
\end{align}
To the best of our knowledge, we are the first to apply the BHC to the discrimination of different (continuous-time) Markov chains. Further, we note that our derivation via variational inference is so far missing in the literature, and we will demonstrate later that it can improve on the classical criterion. 
The VBHC can be related to a popular variational estimator for the mutual information (MI) \cite{Poole2019}, where, however, the marginal likelihood is directly replaced by a variational distribution. From a computational perspective, the (V)BHC corresponds to the following simplification: While computing the EIG (or the MI), the posterior computation over all models has to be performed for each experimental outcome, the (V)BHC only requires a single posterior computation per model. This is especially helpful in settings where (repeated) posterior computation becomes prohibitively expensive. 

\section{Model: Conditional Continuous-time Bayesian Networks}\label{sec:CTBN}
\textbf{Definition.} Analogous to continuous-time Bayesian Networks \cite{Nodelman1995}, we define Conditional Continuous-time Bayesian Networks (cCTBN) as an $N$-component cCTMC over factorizing state-spaces ${\mathcal{S}=\mathcal{X}_1\times\dots\times \mathcal{X}_N}$, evolving jointly as a CTMC given any condition $i$. In this work we consider local interventions, thus we assume $\mathcal{I}=\mathcal{I}_1\times\dots\times \mathcal{I}_N$. We state explicitly that single component states and interventions are entries of the states and interventions of the global cCTMC i.e. $s=(x_1,\dots,x_N)$ for $s\in\mathcal{S}$ with $x_n\in\mathcal{X}_n$ and $i=(i_1,\dots,i_N)$ for $i\in\mathcal{I}$ with $i_n\in\mathcal{I}_n$ for all $n\in\{1,\dots,N\}$. For cCTBNs the parent configuration can be summarized via a directed, but possibly cyclic graph structure $G=(\mathcal{V},\mathcal{E})$ and in general depends on the current intervention. We note that the graph of a cCTBN can be unrolled on an infinitesimal time-grid (with spacing $h$) to reveal the interaction graph $\mathcal{G}$ in time, which is again acyclic ~\cite{Cohn2010,Linzner2018}. The effect of unrolling is illustrated in figure~\ref{fig-conditional-ctbn} a) and b). In this manuscript, our goal is learning the possibly cyclic graph $G$. 

The $n$'th nodes' process $\{X_n(t)\}_{t\geq 0}$ depends on its current state $x_n\in\mathcal{X}_n$, its condition  $i_n\in\mathcal{I}_n$ and of all its parents ${U}_n(t)=u_n$ taking values in $\mathcal{U}^G_n\equiv\bigtimes_{m\in\mathrm{par}^G(n)}\mathcal{X}_m$, with $\bigtimes$ denoting the Cartesian product. For a cCTBN, the global transition matrix $p_h(s'\mid s,i)$ then factorizes over nodes
${p_h(s'\mid s,i)=\prod_{n=1}^N p_h(x'_n\mid x_n,u_n,i_n), }$
into local conditional transition probabilities. We define local transition rates ${{\Lambda}}_n^i:\mathcal{X}_n\times\mathcal{X}_n\times\mathcal{U}_n\rightarrow\mathbb{R}$ for each condition $i\in \mathcal{I}_n$. In the following, we write compactly ${{{\Lambda}}_n^i(x, x',u)\equiv {{\Lambda}}_n(x_n, x'_n,u_n, i_n)}$.
Subsequently, we  can express the local conditional transition probabilities as 
$
{p_h(x'_n\mid x_n,u_n,i_n)=\mathbb{1}(x=x')+{h}{{\Lambda}}_n^i(x', x, u)+o({h})}.
$
Lastly, we mention that an equivalent cCTMC can be constructed by amalgamation~\cite{El-Hay2011}. We can define the global transition rate-matrix $\trans$ through these sets $\trans=\{G,\Lambda\}$. This can be used to solve the Chapman-Kolmogorov equation of a cCTBN by transforming it into an equivalent cCTMC.

\textbf{Properties.}
cCTBNs present an extension to CTBNs as, given any condition $i$, a cCTBN is a CTBN. 
Paths of a CTBN $S^{[0,T]}=\{X_{1}^{[0,T]},\dots,X_{N}^{[0,T]}\}$ assume values in the space of piece-wise constant (cadlag) functions. The path-likelihood of a cCTBN given a condition $i$ is the likelihood of a CTBN \cite{Nodelman2003}
\begin{align}\label{eq:llh-fac}
&p(S^{[0,T]}\mid \Lambda,G,i,s_0)=\\
\nonumber&\prod_{n=1}^{N}p(X_{n}^{[0,T]}\mid X_{\mathrm{par}^G(n)}^{[0,T]},\Lambda_n,i_{n})\mathbb{1}(X_n(0)=x_n^0),
\end{align}
where we introduced the vector-valued path of the cCTMC $S^{[0,T]}=\left\{S(t)\mid 0\leq t \leq T\right\}$ and its components $X_n^{[0,T]}=\left\{X_n(t)\mid 0\leq t \leq T\right\}$ for all $n\in\left\{1,\dots,N\right\}$ and the (vector-valued) initial state of the system $s_0=(x_1^0,\dots,x_N^0)$.
The conditional path-likelihood $p(X_{n}^{[0,T]}\mid X_{\mathrm{par}^G(n)}^{[0,T]},\Lambda_n,i_{n})$  of an individual node $n$ can in turn be expressed in terms of the path-statistics~\cite{Nodelman2003}, the number of transitions of node $n$ from state $x$ to $x'$ given the parents state $u$ denoted by ${M_n(x,x', u)}$ and ${T_n(x, u)}$, which denotes the amount of time node $n$ spent in state $x$
\begin{align}\label{eq:local-llh}
&p(X_{n}^{[0,T]}\mid X_{\mathrm{par}^G(n)}^{[0,T]},\Lambda_n,i_{n})=\\
&\nonumber\prod_{x,x'\neq x,u}\Lambda^i_{n}(x,x',u)^{M_{n}(x,x', u)}e^{-T_{n}(x, u)\Lambda^i_{n}(x,x', u)}.
\end{align}
In \cite{Nodelman2003} it was shown that a marginal likelihood for the structure of a CTBN can be calculated in closed form under the assumption of independent gamma priors over the rates
\begin{align}\label{eq:local-marginal llh}
&p(X_{n}^{[0,T]}\mid X_{\mathrm{par}^G(n)}^{[0,T]},i_n)\propto\\
\nonumber&\prod_{x,x'\neq x,u}\Gamma(\bar{\alpha}_n^i(x,x',u))\bar{\beta}_n^i(x,u)^{-\bar{\alpha}_n^i(x,x',u)},
\end{align}
where $\bar{\alpha}_n^i(x,x',u)=M_{n}(x,x', u)+\alpha_n^i(x,x',u)$ and $\bar{\beta}_n^i(x,u)=T_{n}(x, u)+\beta_n^i(x,u)$ and $\alpha_n^i(x,x',u)$, $\beta_n^i(x,u)$ being the hyper-parameters of the gamma priors.

\textbf{Interventions.} Without loss of generality, we denote $0$ as no intervention. This identifies the model with $i=0$ as the un-intervened on or \emph{original} model.

\emph{Perfect interventions.} A perfect intervention, illustrated in figure~\ref{fig-conditional-ctbn} c), corresponds to Pearls $\mathrm{do}$-operation. While it is known \cite{Pearl2000}, that interventions can be modelled as additional variables, correspondence in the case of cCTBNs can found directly. By setting $\Lambda^i_n(x,x',u)=0$ for the intervened node $n$ with  $i_n\neq 0$, \eqref{eq:local-llh} evaluates to 1 iff  $M_n(x,x',u)=0$ for all $x,x'\in\mathcal{X}_n$ and $u\in \mathcal{U}_n^G$, and $0$ otherwise. Together with the indicator for the initial condition in \eqref{eq:llh-fac} this evaluates to an indicator $\mathbb{1}(X_n^{[0,T]}=x_n^0)$ in the path-likelihood \eqref{eq:llh-fac}. % as in \eqref{eq:do}. 
One then recovers the same relationship as in the static model \eqref{eq:do}
\begin{align*}%\label{eq:llh-fac}
&p(S^{[0,T]}\mid \Lambda,G,i_{\neg k}=0,i_k\neq 0,s_0)=\\
&\nonumber\mathbb{1}(X_k^{[0,T]}=x_k^0)\tilde{p}(X_{\neg k}^{[0,T]}),
\end{align*}
with $\tilde{p}(X_{\neg k}^{[0,T]})\equiv \prod_{n=1,n\neq k}^{N}p(X_{n}^{[0,T]}\mid X_{\mathrm{par}^G(n)}^{[0,T]},\Lambda_n,i_{n})\mathbb{1}(X_n(0)=x_n^0)$ and $i_{\neg k}$ being the vector-valued intervention without the $k$'th component.

\emph{Imperfect interventions}, as studied for example in \cite{Eaton2007}, do not fix the state of the intervened on $n$'th node, but only corresponds to a change to its path-likelihood. In cCTBNs, this is reflected by $\Lambda^{i}_n(x,x',u)$ being an arbitrary function of the condition $i_n\neq 0$. Here the dependency of node $n$ on its parents can but doesn't have to be broken. Imperfect interventions are illustrated in figure~\ref{fig-conditional-ctbn} d).

\section{Experimental Sequence Likelihood of a cCTBN}
We want to calculate the likelihood of a sequence of observations ${\mathcal{H}\equiv \{S^{{[0,T]},k}\mid 0\leq k\leq K\}}$ collected in $K$ different experiments under (possibly) $K$ different interventions ${\Pi\equiv \{i^k\mid 0\leq k\leq K\}}$. 

\textbf{Data-pooling for cCTBNs.}
Given an experimental sequence $\mathcal{H}$ under local interventions $i$, the likelihood of this sequence can be expressed in terms of node-wise likelihoods, independent on the order of the sequence,i.e.
\begin{align}\label{eq:data-pool}
&p(\mathcal{H}\mid\Lambda,G,\Pi)=\prod_{n=1}^{N}\prod_{i\in \mathcal{I}_n}\prod_{x,x'\neq x,u}\\
\nonumber&\Lambda^i_{n}(x,x', u)^{\bar{M}_{n}(x,x', u,i)}\exp\left[-\bar{T}_{n}(x, u)\Lambda^i_{n}(x,x', u))\right],
\end{align}
with sufficient statistics $\bar{T}_{n}(x, u,j)=\sum_{k=1}^{K}\mathbb{1}(i^k_{n}=j)T^k_{n}(x, u)$ and $\bar{M}_{n}(x,x', u,j)=\sum_{k=1}^{K}\mathbb{1}(i^k_{n}=j)M^k_{n}(x,x', u)$.
This can be directly seen by considering the joint likelihood 
\begin{align}
\nonumber p(\mathcal{H}\mid  \Lambda,G,\Pi)=\prod_{k=1}^K\prod_{n=1}^{N}p(X_{n}^{[0,T],k}\mid X_{\mathrm{par}^G(n)}^{[0,T],k},\Lambda_n,i_{n}^k).
\end{align}
Inserting the likelihood \eqref{eq:local-llh} of $X_{n}^{[0,T],k}\mid X_{\mathrm{par}(n)}^{[0,T],k},\Lambda_n,i_{n}^k$, the above claim follows after definition of the statistics.

The above result allows us to use samples generated under interventions for the estimation of the original CTBN with $i=0$, if each node was observed under condition $i=0$ at least once. It reveals the penalty of performing an experiment under a perfect intervention: The statistics of intervened-on node remain unchanged $-$ the node is unobserved. We give a simple example that interventional data can be more informative than complete observations in the presence of time-scale separation.

\textbf{Example: Interventional Data can be more informative than complete observations.}
Consider a minimal example of a two node CTBN $X\rightarrow Y$ with time-scale separation. For simplicity, we assume $\Lambda_X(x,x')=\varepsilon\rightarrow 0$, but the argument translates to finite rates. Further, we assume $\Lambda_X(x',x)> 0$ and $X(0)=x$. Then $X(t)=x'$ can not be observed for any finite $t\geq 0$ and the posterior over any rate $\Lambda_Y(y,y', x')>0$ remains the prior and can thus not be learned. To see this we compute the posterior
\begin{align*}
&p(\Lambda_Y(y,y', x')\mid Y^{[0,T]},X^{[0,T]}=x)=\\
\nonumber&\frac{p( Y^{[0,T]}\mid X^{[0,T]}=x,\Lambda_Y(y,y', x'))p(\Lambda_Y(y,y', x'))}{p( Y^{[0,T]}\mid X^{[0,T]}=x)},
\end{align*}
however, by checking the likelihood \eqref{eq:local-llh}, we find $p( Y^{[0,T]}\mid X^{[0,T]}=x, \Lambda_Y(y,y', x'))=p( Y^{[0,T]}\mid X^{[0,T]}=x)$, because the number of transitions $M_Y(y,y', x')=0$, thus ${p(\Lambda_Y(y,y', x')\mid Y^{[0,T]},X^{[0,T]}=x)=p(\Lambda_Y(y,y', x'))}$ and no information has been gained.
On the other hand, an intervention allows us to set $X^{[0,T]}=x'$, thus, for the observation time $T$ being sufficiently large, we will have $M_Y(y,y', x')>0$ almost surely and
$
p( Y^{[0,T]}\mid\mathrm{do}(X^{[0,T]}=x'), \Lambda_Y(y,y', x'))\neq p( Y^{[0,T]}\mid \mathrm{do}(X^{[0,T]}=x')).
$
Thus we have a finite information gain. 
\begin{figure*}[t]
\begin{centering}
        \includegraphics[width=1.99\columnwidth]{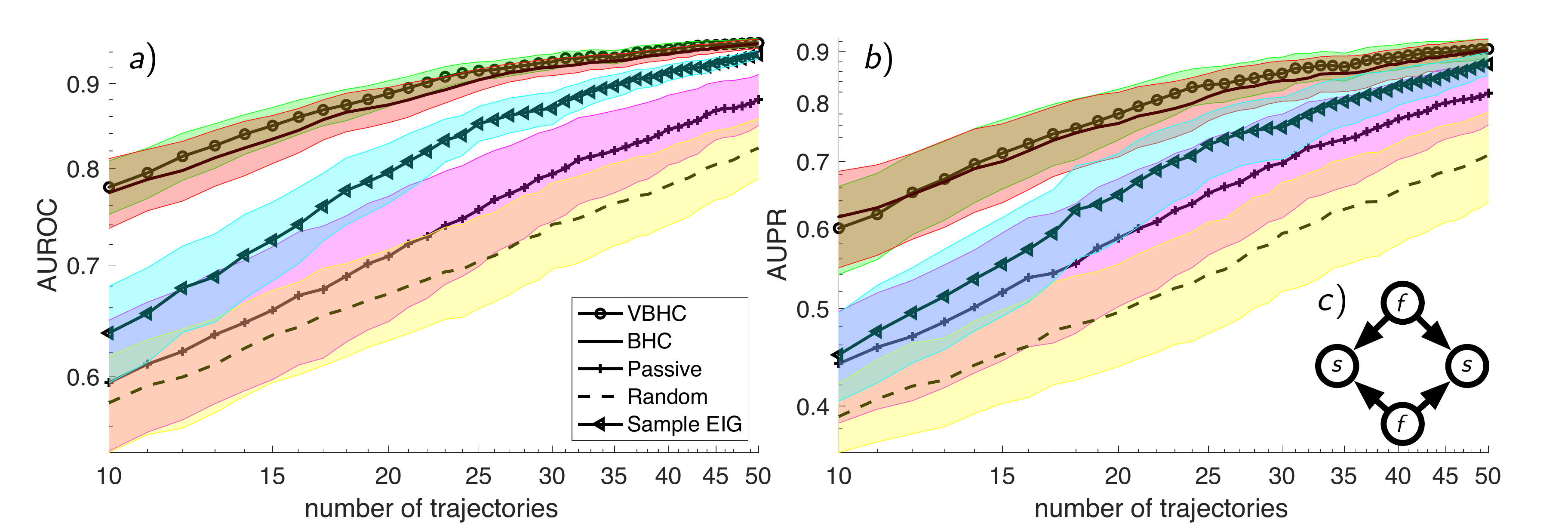}
    \caption{Results of active structure learning from synthetic data. We plotted mean (line) and variance (areas) after repeating structure learning for $500$-times for each design. We plotted a) AUROC and b) AUPR for a different number of sequential experiments. In c) we denoted the ground truth graph, where f denotes fast and s slow nodes.}
    \label{fig-structure}
    \end{centering}
\end{figure*}
\section{Active Learning of CTBNs}
 In the following, we will use (V)BHC for choosing sequences of interventions for repeated experiments. To facilitate this, we now derive semi-analytical expressions for the (V)BHC. We restrict ourselves to perfect interventions only. To avoid clutter, we define the set of nodes not subject to an intervention $\aleph\equiv\left\{n\in\left\{1,\dots,N\right\}\mid i_n=0\right\}$. In the following, experimental outcomes are possible paths of a cCTBN $D=S^{[0,T]}$ and $\Lambda^{0}\equiv \Lambda^{i=0}$ are the rates of the original model. As for the computation of the (V)BHC we marginalize over possible future outcomes, we introduce the notation $\hat{S}^{[0,T]}$ for a possible outcome path ${\hat{S}^{[0,T]}\sim p(S^{[0,T]}\mid \Lambda,G,i,s_0)}$ and its statistics, the number of transitions $\hat{M}_{n}(x,x',u)$ and dwell times $\hat{T}_{n}(x,u)$ of the $n$'th node for all $x,x'\in\mathcal{X}_n$ and $u \in\mathcal{U}_n$.
\begin{figure*}[t]
\begin{centering}
   \includegraphics[width=1.9\columnwidth]{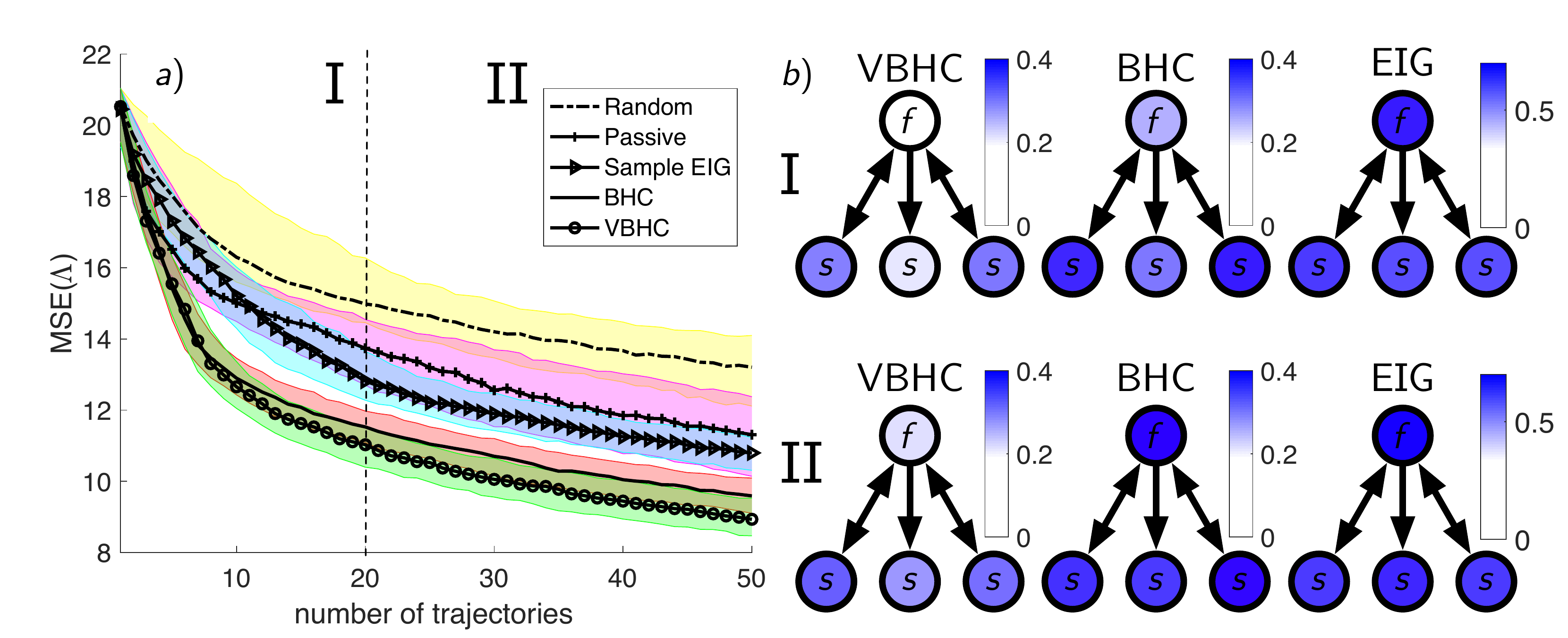}
    \caption{a) Results of active parameter learning in from synthetic data. We plotted mean (line) and 25-75$\%$ confidence intervals of the mean-squared error of the estimated rates for various experimental designs after repeating parameter learning for $500$ times. In b) we denote the graph, where f denotes fast and s slow nodes. Color-scales denote the marginal probability of intervening on this node in the $k$'th experiment for regions I and II, see a).
    }
    \label{fig-rate}
    \end{centering}
\end{figure*}

\textbf{Active Parameter Learning.}
The VBHC can be calculated as an expected KL-divergence between different models. In the following, we derive the criterion for optimal rate-estimation, given the structure $G$. The KL-divergence between two cCTBNs, with different rate-matrices $\Lambda$ and $\Lambda'$ under the same intervention $i$ and same graph $G$ read 
\begin{align*}
&\KL{p(S^{[0,T]}\mid \Lambda,G,i)}{p(S^{[0,T]}\mid \Lambda',G,i)}=\\
&\sum_{n\in\aleph}\sum_{x,x',u}\{\mathsf{{E}}[\hat{M}_{n}(x,x', u)\mid \Lambda,G,i]\ln\frac{\Lambda^0_{n}(x,x', u)}{\Lambda^{'0}_{n}(x,x', u)}\\
&\nonumber-\mathsf{{E}}[\hat{T}_{n}(x, u)\mid \Lambda,G,i]\{ \Lambda^0_{n}(x,x', u)-\Lambda^{'0}_{n}(x,x', u)\}\}. 
\end{align*} 
The expected path-statistics of possible outcomes $\hat{M}_{n}$ and $\hat{T}_{n}$, can be calculated analytically and are given by the solution $p_t(s\mid s_0,i)$ of the Chapman-Kolmogorov equation \eqref{eq:m-eq} 
\begin{align}\label{eq:moments-time}
&\mathsf{{E}}[\hat{T}_{n}(x, u)\mid  \Lambda,G,i]\quad\,\,\,\,=\\
&\nonumber\sum_{s}\mathbb{1}(s_n=x)\mathbb{1}(s_{\mathrm{par}^G(n)}=u)\int_0^T\mathrm{d}t\,p_t(s\mid s_0,i),\\\label{eq:moments-trans}
&\mathsf{{E}}[\hat{M}_{n}(x,x', u)\mid  \Lambda,G,i]=\\
&\nonumber\mathsf{{E}}[\hat{T}_{n}(x, u)\mid  \Lambda,G,i]\Lambda^0_n(x,x', u).
\end{align}
While the exact solution of the master-equation is a limiting factor for larger graphs, we note that scalable methods that approximate it exist~\cite{Cohn2010,El-Hay2010,Linzner2018}. For a derivation of the expected moments, see Appendix B.2, for a derivation of the KL-divergence B.1. 

Under a Gamma prior for $\Lambda$, the posterior $p(\Lambda\mid\mathcal{H},G)$, calculated using \eqref{eq:data-pool}, will be a Gamma distribution due to conjugacy $p(\Lambda\mid\mathcal{H},G)=\prod_{n,x,x',u}\mathrm{Gam}\left(\Lambda_n(x,x', u)\mid \bar{\alpha}_n(x,x', u),\bar{\beta}_n(x,u)\right)$, with $\bar{\alpha}_n(x,x', u)=\bar{M}_{n}(x,x', u,i=0)+\alpha_n(x,x',u)$ the posterior transition counts and $\bar{\beta}_n(x,u)={\bar{T}_{n}(x,u,i=0)+\beta_n(x, u)}$ the posterior waiting times. Consequently, we choose the same parametric form for $q_\kappa(\Lambda)=\prod_{n,x,x',u}\mathrm{Gam}\left(\Lambda_n(x,x', u)\mid \alpha^{\kappa}_n(x,x',u),\beta^{\kappa}_n(x, u)\right)$, with $\alpha^{\kappa}_n(x,x', u)$ and $\beta^{\kappa}_n(x, u)$ being variational parameters. We finally arrive at the semi-analytical expression for the VBHC in terms of the variational parameters and the expected path-statistics
\begin{align}\label{eq:vbhc_params}
&\mathrm{VBHC}(i,\kappa)=\KL{q_\kappa(\Lambda)}{p(\Lambda\mid\mathcal{H},G)}\\
&\nonumber-\int d\Lambda\:p(\Lambda\mid\mathcal{H},G)\sum_{n\in \aleph}\sum_{x,x',u}\\
\nonumber&\times\{\mathsf{{E}}[\hat{M}_{n}(x,x', u)\mid  \Lambda,G,i]
\\&\times(\ln \Lambda-\psi^{(0)}(\alpha^{\kappa}_n(x,x', u))+\ln\beta^{\kappa}_n(x, u))\nonumber\\
&\nonumber-\mathsf{{E}}[\hat{T}_{n}(x, u)\mid  \Lambda,G,i]\left( \Lambda-\frac{\alpha^{\kappa}_n(x,x', u)}{\beta^{\kappa}_n(x, u)}\right)\bigg\},
\end{align}
where $\psi^{(k)}$ is the di-gamma function of $k$'th order. The KL-divergence $\KL{q_\kappa(\Lambda)}{p(\Lambda)}$ between two Gamma distributions has a closed form expression, -- see Appendix B.3. We can compute the gradients $\partial_{\kappa}\mathrm{VBHC}(i,\kappa)$ analytically, which facilitates optimization of the bound for large systems.
\newline\newline
\textbf{Active Structure Learning.} The KL-divergence between two marginal CTBNs reads
\begin{align*}
&\KL{p(S^{[0,T]}\mid G,i)}{p(S^{[0,T]}\mid G',i)}=\\
&\nonumber\sum_{n\in\aleph}\sum_{x,x'}\sum_{u\in \mathcal{U}_n^G}\sum_{u'\in \mathcal{U}_n^{G'}}\expectation{}{\mathsf{{E}}\bigg[\ln\frac{\Gamma(\tilde{\alpha}_{n}(x,x', u\,))}{\Gamma(\tilde{\alpha}_{n}(x,x', u'))}\bigg|\,\Lambda,G,i \bigg]}\\
&+\expectation{}{\mathsf{{E}}\bigg[\tilde{\alpha}_{n}(x,x',u')\ln\tilde{\beta}_{n}(x, u')\bigg|\,\Lambda,G,i \bigg]}\\
&-\expectation{}{\mathsf{{E}}\bigg[\tilde{\alpha}_{n}(x,x',u\,)\ln\tilde{\beta}_{n}(x, u\,)\,\bigg|\,\Lambda,G,i \bigg]},
\end{align*}
with the outer expectation w.r.t $p(\Lambda\mid \mathcal{H},G)$ and the inner one w.r.t the path-likelihood ${p(S^{[0,T]}\mid \Lambda,G,i,s_0)}$. Further we defined short-hands $\tilde{\alpha}_{n}(x,x',u')\equiv \hat{M}_{n}(x,x', u\,)+\alpha_{n}(x,x', u\,)$ and $\tilde{\beta}_{n}(x,u\,)\equiv \hat{T}_{n}(x,u)+\beta_{n}(x, u)$.
% up to second order. As this KL-expansion is quite lengthy, see Appendix B.5 for its exact form and derivation. There, we also give derivation of closed form expressions of the non-zero higher moments $\mathsf{{E}}[\hat{M}_{n}(x,x', u\,)^2]$, $\mathsf{{E}}[\hat{T}_{n}(x, u\,)^2]$ and $\mathsf{{E}}[\hat{M}_{n}(x,x', u\,)\hat{T}_{n}(x, u\,)]$ in terms of the solution of the master-equation, which we were unable to find in literature. 
The posterior over structures can be computed in closed form, see \eqref{eq:local-marginal llh}, by marginalization $p(G\mid \mathcal{H})=\int \mathrm{d}\Lambda\,p(\Lambda\mid \mathcal{H},G)p(G)$, with any pior $p(G)$ (we assume a uniform categorical). It is natural to assume a categorical distribution $q_\kappa(G)=\mathrm{Cat}(\kappa_G)$.Then the VBHC for structure learning reads
\begin{align}\label{eq:vbhc_struct}
&\mathrm{VBHC}(i,\kappa)=\KL{q_\kappa(G)}{p(G\mid\mathcal{H} )}\\
&\nonumber-\mathsf{E}[\mathsf{E}[\mathrm{KL}(p(S^{[0,T]}\mid G,i)\,||\,p(S^{[0,T]}\mid G',i)]],
\end{align}
with the inner expectation w.r.t $q_\kappa(G)$, the outer w.r.t $p(G'\mid \mathcal{H})$.
As the moments ${\mathsf{{E}}[\ln \hat{M}_{n}\mid \Lambda,G,i]}$ and ${\mathsf{{E}}[\ln \hat{T}_{n}\mid\Lambda,G,i ]}$ can not be evaluated in closed form, we approximate the KL-divergence by a first-order expansion around the moments \eqref{eq:moments-time} and \eqref{eq:moments-trans}, see Appendix B.4 for details.
\newline\newline
\textbf{Discussion.} We emphasize, that the expressions \eqref{eq:vbhc_params} and  \eqref{eq:vbhc_struct}, enable us to perform active parameter and structure learning in high-dimensions, as the integration over outcomes is performed analytically. As mentioned in section~\ref{section-VBHC}, the corresponding BHC is computed by setting $q_\kappa(\Lambda)=p(\Lambda\mid \mathcal{H},G)$ in \eqref{eq:vbhc_params} and  $q_\kappa(G)=p(G\mid \mathcal{H})$ in \eqref{eq:vbhc_struct}. For the (V)BHC for parameter- and structure learning, only the integral over ${p(\Lambda\mid \mathcal{H},G)}$ can not be evaluated analytically, as for each realization of $\Lambda$, the full master-equation needs to be solved. We thus approximate this integral by $N_S$ posterior samples. We summarized the computational steps mentioned above in algorithmic format in Appendix A. 

To highlight the computational benefit of the (V)BHC, we compare inference complexity in the number of nodes $N$ and the cardinality of local state spaces $|\mathcal{X}|$.
For the EIG, one needs to integrate the posterior over parameters for all possible paths the network can take. The
complexity here is two-fold: \begin{inlineitemize}\item  The number of possible paths scales as $|\mathcal{X}|^N$ and \item the complexity of a posterior-update,
which is dominated by computing conditional summary statistics for transitions is $N|\mathcal{X}|^{N}$.
\end{inlineitemize}
Thus, in total the complexity
of calculating the EIG is given by the product $N|\mathcal{X}|^{2N}$. For the (V)BHC calculation of the conditional summary statistics
is only performed once. Thus, we have a sum instead of a product $N|\mathcal{X}|^{N} + |\mathcal{X}|^N$.

\section{Experiments}
We evaluate the performance of interventions selected by different designs for sequential experiments. In all considered scenarios, interventions are tuples of node and state pairs, $i=((m,x_m),(j,x_j),\dots)$ with $m,j\in\{1,\dots,N\}$ and $x_m\in \mathcal{X}_m,x_j\in\mathcal{X}_j$, which corresponds to a $\mathrm{do}$-operation of setting the $m$'th node into state $x_m$ and the $j$'th node into state $x_j$. In the following, we search the optimal intervention in the set of all possible interventions of this type.
In order to select the optimal invervention, we compare the EIG~\eqref{eq:EIG} and the (V)BHC~\eqref{eq:int-VBHC} given \eqref{eq:vbhc_struct} for structure and \eqref{eq:vbhc_params} for parameter learning, and also compare to random or no interventions (passive). Neither the EIG, nor the (V)BHC can be computed completely in analytical form. We employ a (nested) monte-carlo scheme for the computation of the EIG, see Appendix A. For the computation of the (V)BHC, we approximate the integration over rates via posterior sampling. We discuss using sample estimates of design criteria in Appendix C.1. For the problem sizes considered, the (exact) solution of the master-equation has a negligible contribution to the run-times, compared to the repeated posterior computations. Thus, we throughout compare EIG and (V)BHC given the same number of posterior samples $N_S$. Minimization of the VBHC is feasible using standard Matlab optimizers as gradients $\partial_\kappa \mathrm{VBHC}(i,\kappa)$ are computed analytically.

\begin{figure}[t]
\begin{centering}
\includegraphics[width=0.49\textwidth]{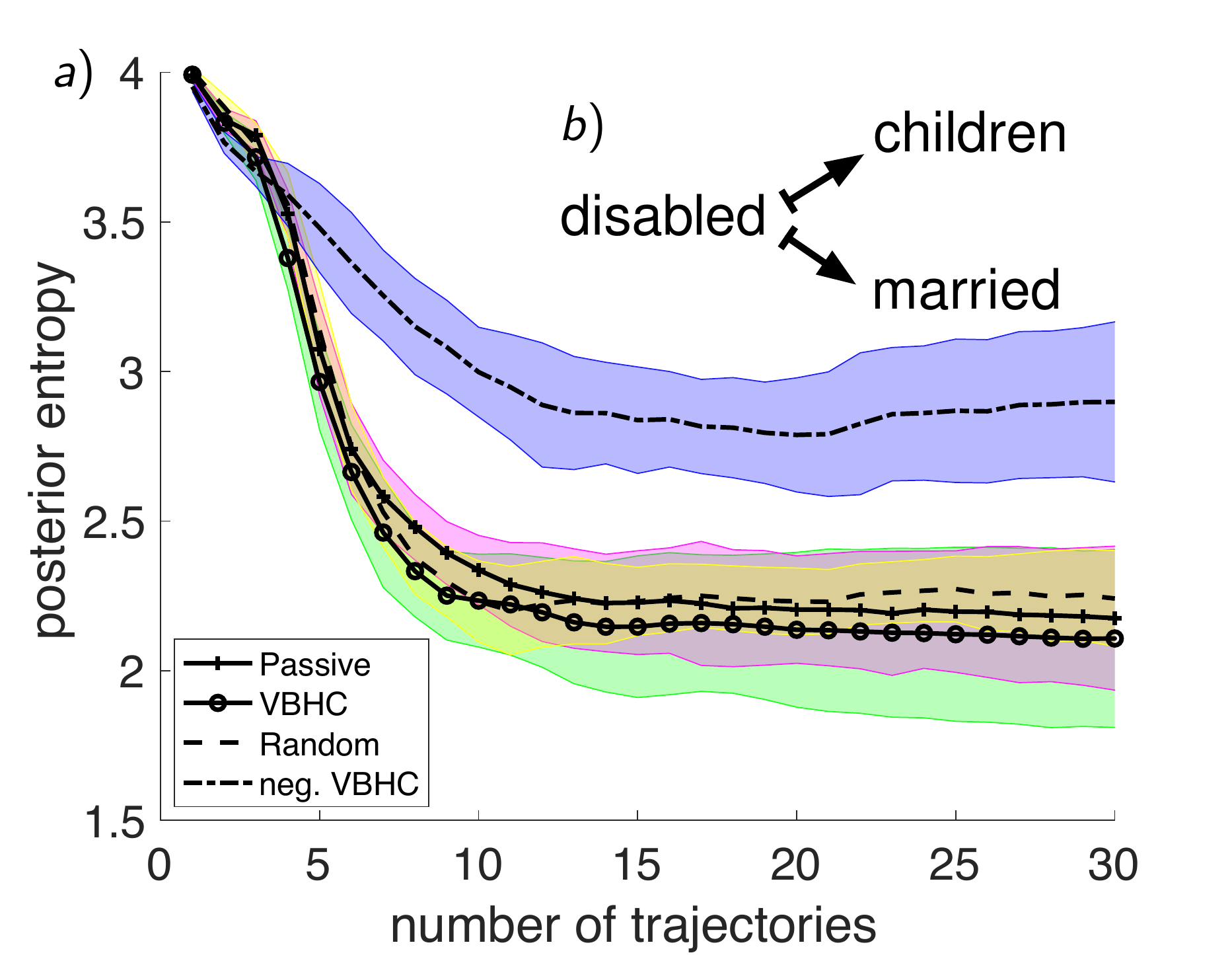}
  \caption{a) Mean and variance (area) of the evolution of the posterior entropy in BHPS data-set for 100 repetitions. b) Sketch of the underlying network, where "disabled" is a root node.}
   \label{fig-bhps}
    \end{centering}
\end{figure}
\subsection{Synthetic Data}
For synthetic experiments, we learn structures and parameters from randomly generated CTBNs with $L=4$ nodes with binary state-spaces. As the amount of data needed to identify structures and parameters varies strongly with the underlying ground truth, we display results for fixed ground truths. We generate problems, which contain features that can be identified and exploited by an active learning scheme. As shown in this manuscript, one way to do this is by inducing time-scale separation. For structure and parameter learning, we thus create problems containing fast and slow nodes. In both cases, we exhaustively search in the space of all possible interventions, including targeting multiple nodes or performing no interventions. For each trajectory, an initial state $s_0$ is drawn at random, which is accessible to the design (although a distribution of initial states could also be learned). We set $N_S=10$ and the length of each trajectory fixed to be $\tau=3$ a.u..

\textbf{Active Structure Learning.}
We generate ground truth rates by sampling rate-parameters from independent Gamma-distributions ${\Lambda^*\sim\mathrm{Gam}(\Lambda\mid \alpha_{f/s},\beta_{f/s})}$ with two different types hyper-parameters $f$ and $s$, referring to fast and slow nodes, respectively. We used $\alpha_f=5$, $\alpha_s=1/5$ and $\beta_{f}=\beta_{f}=1$. We then perform structure learning via exhaustive scoring as in \cite{Nodelman2003}. 
As a metric for structure learning, we employ the area under the Receiver-Operator-Characteristic curve (AUROC) and Precision-Recall curve (AUPR), which are frequently used to quantify the performance of (structure) classifiers. For an unbiased classifier, both metrics should approach one in the limit of infinite data. In figure~\ref{fig-structure} a) and b), respectively, we show the results for structure learning. As AUROC and AUPR can only be calculated w.r.t a ground truth, they can not be made an objective for a design. 

\textbf{Active Parameter Learning.}
For parameter learning, we fix the ground truth structure and the ground truth rates across experiments. Following CTBN literature \cite{Cohn2010,El-Hay2011}, we chose the ground truth rates $\Lambda^*$ as scaled softmax functions $\Lambda^*_n(x,x', u)=r_{f/s}\mathrm{softmax}(\gamma\sum_{y\in u}\mathbb{1}(x=y))$ for the $n$'th node, with $\gamma=3$ and $r_s=1/5$ and $r_f=5$ for slow and fast nodes, respectively.
We quantify the performance of parameter learning by the posterior averaged mean-squared-error (MSE), defined by $\mathrm{MSE}(\Lambda)= \mathsf{E}[\left(\Lambda-\Lambda^*\right)^2]$, with the expectation subject to the current rate-posterior $p(\Lambda\mid \mathcal{H},G)$. The results of this experiment are shown in figure~\ref{fig-rate}~a). Similar to the case of structure learning, the ground truth rate-matrix is unavailable during the experiment, and the MSE can thus not be used as a design objective. We also investigate the qualitative behavior of the designs in figure~\ref{fig-rate}~b), where we provide the marginal probability of intervening on a node in the $k$'th step of the experiment. For the (V)BHC it can be seen, that recommended interventions follow the intuition of exploiting time-scale separation, as interventions targeting slow nodes pointing on the fast node have higher probability.

%\begin{wrapfigure}{h}{0.5\textwidth}
%  \begin{center}
%    \includegraphics[width=0.44\textwidth]{MI_design}
%  \end{center}
%  \caption{Mutual information between design sample estimates and recommended interventions for different number of samples $N_S$. Areas denote 25-75$\%$ confidence intervals. }
%   \label{fig:MI_design}
%\end{wrapfigure}

%\textbf{Sample estimates of Design Criteria.}
%We want to investigate the viability of using sample estimates of different criteria for active learning of CTBNs. One basic requirement on such an estimate is that its recommendations actually depend on the history of observations  $\mathcal{H}\underset{\mathrm{design}}{\longrightarrow}i$. We can make this formal by the following non-parametric dependency check: The recommended intervention $i$ is dependent on experimental sequence $\mathcal{H}$ if they share high mutual information $I(i,\mathcal{H})$. We stress, that this does not reflect the quality of recommended interventions! We calculate the MI for random graphs of size $L=3$ for different sample sizes $N_S$ for random histories $\mathcal{H}$ consisting of 30 trajectories drawn from our synthetic network. The results are displayed in figure \ref{fig:MI_design}. For all sample sizes considered, (V)BHC shares a much higher MI with their recommended interventions, than the sample estimate of the EIG.
\subsection{Real-World Data}
\textbf{British Household Data-set.}
We apply our method to the British Household Panel Survey (BHPS)~\cite{bhps}. This data-set has been collected yearly from 1991 to 2002, thus consisting of 11 time-points. Each of the 1535 participants was questioned about several facts (variables) of their life. We choose 3 variables "marital status", "has children under 12" and "is registered disabled" of those facts. We had no means of intervening on this data-set directly, but can use it as a proof of concept in order to verify our method on real world data. For this, we identified "is registered disabled" as a root node, meaning there is no edge from "marital status" or "has children under 12" to this node. We confirmed this by performing network inference on the full data-set, see inlet \ref{fig-bhps} b). We then selected for a subset where the variable "is registered disabled" remains in one state during the complete trajectory. We interpreted these cases as interventions, which is valid as conditioning on a root node is the same as intervening on a root node. We then simulated an experiment, where one either draws a data-point from the full-set or from the subset of interventions on "is registered disabled". We performed this experiment for passive, random, VBHC designs and negative VBHC design (always pick the worst possible intervention) for structure learning with $N_S=40$ posterior samples as recommenders for interventions. For details on the processing of this (incomplete) data-set, we refer to the Appendix C.2. As for the real-world data-set no ground truth structure is available, we track the evolution of the posterior entropy over structures for 100 independent runs, see figure~\ref{fig-bhps} a). In Appendix C.3, we show that for all designs the inferred network converge against the one inferred using the full data-set (using AUROC and AUPR as metrics). We note that the effect of active learning can be expected to be small in this synthetic scenario, as we were only able to intervene on a single node.

\section{Conclusion}
We presented a novel variational criterion for active learning, that alleviates the curse of dimensionality when integrating over all possible experimental outcomes. We presented cCTBNs, a framework to study the effect of interventions on CTBNs. We have shown that our novel criterion can be calculated semi-analytically for cCTBNs and can be used to recommend interventions that speed-up structure and parameter learning on synthetic and real-world data. In this manuscript, we performed exact inference and exhaustive search, limiting us to small graphs. However, many principled approximation techniques for inference~\cite{Cohn2010,El-Hay2010,El-Hay2011,Rao2012}, and structure learning~\cite{Nodelman2005,Linzner2019} are available that allow learning of large scale CTBNs and are compatible with our framework. 

\section*{Acknowledgements}
We thank the anonymous reviewers for helpful comments on the previous version of this manuscript.
D.~L. and H.~K. acknowledge funding by the European Union's Horizon 2020 research and innovation programme (iPC--Pediatric Cure, No. 826121). H.~K. acknowledges support by the Hessian research priority programme LOEWE within the project CompuGene.

\bibliography{cv_learning_ctbn}

\appendix

\onecolumn
\icmltitle{Supplement: Active Learning of Continuous-time Bayesian Networks through \\Interventions}



% Supplementary material: To improve readability, you must use a single-column format for the supplementary material.

 \section{Algorithms}
 
  \begin{algorithm}[h!]
   \caption{Computation of VBHC for Parameter Learning}
   \label{alg:active-learning}
\begin{algorithmic}[h1]
   \STATE {\bfseries Input}: Proposed intervention $i$, current initial state $s_0$, desired number of posterior-samples $N_S$, current belief over parameters $p(\Lambda\mid\mathcal{H},G)$, initial variational parameters $\boldsymbol{\alpha}^{\kappa,0}$, $\boldsymbol{\beta}^{\kappa,0}$. As initial values for the optimization, we set the initial values for the optimization to the posterior counts $\boldsymbol{\alpha}^{\kappa,0}=\bar{\boldsymbol{\alpha}}$, $\boldsymbol{\beta}^{\kappa,0}=\bar{\boldsymbol{\beta}}$, see main-text.
   \FOR{$n_S$ = 1 : $N_S$}
        \STATE Draw $\hat{\Lambda}_{n_S}\sim p(\Lambda\mid\mathcal{H},G)$.
        \STATE Perform intervention by setting $\hat{\Lambda}_{{n_S},n}=0$ and initial state $x_n^0$ for all $n\notin\aleph$. 
        \STATE Calculate $W$ from $\hat{\Lambda}_{n_S}$ and $G$ by amalgamation.
        \STATE Solve the master-equation main-text (3) subject to $W$ and $s_0$ and recover $\mathsf{E}[\hat{M}(s,s')]$ and $\mathsf{E}[\hat{T}(s)]$ using appendix \eqref{eq:expected-trans} and \eqref{eq:expected-dwell} respectively.
        \FOR{$n$ = 1 : $N$}
              \STATE Compute expected statistics $\mathsf{E}[\hat{M}_n\mid \hat{\Lambda}_{n_S},G,i]$ and $\mathsf{E}[\hat{T}_n\mid \hat{\Lambda}_{n_S},G,i]$ from $\mathsf{E}[\hat{M}(s,s')]$ and $\mathsf{E}[\hat{T}(s)]$, see appendix \eqref{eq:projection-trans} and \eqref{eq:projection-dwell}.
        \ENDFOR
   \ENDFOR
   \STATE Calculate $\mathrm{VBHC}(i,\kappa)$ via main-text (12) and gradients appendix \eqref{eq:gradients-alpha} and \eqref{eq:gradients-beta} with weighted posterior samples replacing ${\int \mathrm{d}p(\Lambda\mid\mathcal{H},G)}$ with ${\sum_{n_S=1}^{N_S} p(\hat{\Lambda}_{n_S}\mid\mathcal{H},G)}$.
   \STATE Minimize w.r.t $\kappa$.
   \STATE {\bfseries Output}: $\min_{\kappa}\mathrm{VBHC}(i,\kappa)$.
\end{algorithmic}
\end{algorithm}

  \begin{algorithm}[!h]
   \caption{Computation of VBHC for Structure Learning}
   \label{alg:active-learning-structure}
\begin{algorithmic}[1]
   \STATE {\bfseries Input}: Proposed intervention $i$, current initial state $s_0$, desired number of posterior-samples $N_S$, current belief over parameters $p(\Lambda\mid\mathcal{H},G)$ and structures $p(G\mid\mathcal{H})$, initial variational parameters $\kappa$.
   \FOR{$n$ = 1 : $N$}
         \FOR{$n_S$ = 1 : $N_S$}
             \FOR{$\mathrm{par}(n)$ in power-set of $\mathcal{V}$}
                 \STATE Draw $\hat{G}_{n_S}\sim p(G\mid\mathcal{H},\mathrm{par}(n))$.
                 \STATE Draw $\hat{\Lambda}_{n_S}\sim p(\Lambda\mid\mathcal{H},\hat{G}_{n_S})$.
                 \STATE Perform intervention by setting $\hat{\Lambda}_{{n_S},n}=0$ and initial state $x_n^0$ for all $n\notin\aleph$. 
                 \STATE Calculate $W$ from $\hat{\Lambda}_s$ and $\hat{G}_{n_S}$ by amalgamation.
                 \STATE Solve the master-equation main-text (3) subject to $W$ and recover $\mathsf{E}[\hat{M}(s,s')]$ and $\mathsf{E}[\hat{T}(s)]$.
       
                \STATE Compute expected statistics $\mathsf{E}[\hat{M}_n\mid \hat{\Lambda}_{n_S},\hat{G}_{n_S},i]$ and $\mathsf{E}[\hat{T}_n\mid \hat{\Lambda}_{n_S},\hat{G}_{n_S},i]$ from $\mathsf{E}[\hat{M}(s,s')]$ and $\mathsf{E}[\hat{T}(s)]$, see appendix \eqref{eq:projection-trans} and \eqref{eq:projection-dwell}.
             \ENDFOR
        \ENDFOR
   \ENDFOR
   \STATE Calculate $\mathrm{VBHC}(i,\kappa)$ using appendix~\eqref{eq:VBHC-struct-sample} and gradients appendix~\eqref{eq:grad-VBHC-struct-sample} with weighted posterior samples replacing $\sum_{G\mid \mathrm{par}(n)}
\sum_{G'\mid \mathrm{par}'(n)} 
p(G\mid \mathcal{H})q(G')\int\mathrm{d}p(\Lambda\mid G)$ with ${\sum_{n_S,n'_S}p(\hat{G}_{n_S}\mid \mathcal{H})q(\hat{G}_{n'_S})p(\hat{\Lambda}_{n_S}\mid \hat{G}_{n_S})}$.
   \STATE Minimize w.r.t $\kappa$.
   \STATE {\bfseries Output}: $\min_{\kappa}\mathrm{VBHC}(i,\kappa)$.
\end{algorithmic}
\end{algorithm}

\begin{algorithm}[!]
\caption{Computation of the EIG for Parameter Learning}
\label{alg:eig-active-learning}
\begin{algorithmic}[1]
   \STATE {\bfseries Input}: Proposed intervention $i$, current initial state $s_0$, desired number of posterior-samples $N_S$, number of path samples $N_P$, current belief over parameters $p(\Lambda\mid\mathcal{H},G)$.
   \STATE Set $\mathrm{EIG}=0$.
   \FOR{$n_S$ = 1 : $N_S$}
        \STATE Draw parameter $\hat{\Lambda}_{n_S}\sim p(\Lambda\mid\mathcal{H},G)$
        \STATE Perform intervention by setting $\hat{\Lambda}_{{n_S},n}=0$ and initial state $x_n^0$ for all $n\notin\aleph$.  
   \FOR{$n_p$ = 1 : $N_P$}
       \STATE Draw path ${\hat{S}^{[0,T]}\sim p(S^{[0,T]}\mid \hat{\Lambda}_{n_S},G,i,s_0)}$.
         \STATE Set $\mathrm{EIG} = \mathrm{EIG} + \frac{1}{N_P N_S} \left( \ln p(\hat{\Lambda}_{n_S}\mid\hat{S}^{[0,T]},\mathcal{H})-\ln p(\hat{\Lambda}_{n_S}\mid \mathcal{H})\right)$.
   \ENDFOR
   
   \ENDFOR
   \STATE {\bfseries Output: } Estimate EIG.
\end{algorithmic}
\end{algorithm}

 \begin{algorithm}[!]
\caption{Computation of the EIG for Structure Learning}
\label{alg:eig-active-learning-structure}
\begin{algorithmic}[1]
   \STATE {\bfseries Input}: Proposed intervention $i$, current initial state $s_0$, number of path samples $N_S$, current belief over parameters $p(\Lambda\mid\mathcal{H},G)$ and structures $p(G\mid\mathcal{H})$.
   \STATE Set $\mathrm{EIG}=0$.
   \FOR{$n$ = 1 : $N$}
      \FOR{$\mathrm{par}(n)$ in power-set of $\mathcal{V}$}
             \FOR{$n_s$ = 1 : $N_S$}
                 \STATE Draw $\hat{G}_{n_S}\sim p(G\mid\mathcal{H},\mathrm{par}(n))$.
                \STATE Draw parameter $\hat{\Lambda}_{n_S}\sim p(\Lambda\mid\mathcal{H},\hat{G}_{n_S})$.
              \STATE Perform intervention by setting $\hat{\Lambda}_{{n_S},n}=0$ and initial state $x_n^0$ for all $n\notin\aleph$.  
               \STATE Draw path ${\hat{S}^{[0,T]}\sim p(S^{[0,T]}\mid \hat{\Lambda}_{n_S},\hat{G}_{n_S},i,s_0)}$.
             \STATE Set $\mathrm{EIG} = \mathrm{EIG} + \frac{1}{N_S} \left( \ln p(\mathrm{par}(n)\mid \mathcal{H},\hat{S}^{[0,T]})-\ln p(\mathrm{par}(n)\mid \mathcal{H})\right)$,\\ see appendix~\eqref{eq:par-post}.
              \ENDFOR
      \ENDFOR
   \ENDFOR
   \STATE {\bfseries Output: } Estimate EIG.
\end{algorithmic}
\end{algorithm}
\section{Derivations}
All derivations are done for a fixed set of conditions $i$ and respective initial states $s_0$. We will omit those in the following derivations for readability.
\subsection{Kullback--Leibler divergence between two CTBNs}
Evaluation of our design criteria (V)BHC, requires the calculation of the KL-divergence between two cCTBNs. 

The likelihood of observing a CTBN path $D=S^{[0,T]}$ is (expressed in terms of its sufficient statistics)
\begin{align}\label{eq:llh-ctbn}
p(S^{[0,T]}\mid\Lambda,G)=\prod_{n}\prod_{x,x',u}\Lambda_{n}(x,x', u)^{{M}_{n}(x,x',u)}\exp{\left\{ -\Lambda_{n}(x,x',u){T}_{n}(x, u)\right\} }.
\end{align}
The KL between two measures is defined via the integration over all paths
\begin{align*}
\KL{p(S^{[0,T]}\mid\Lambda,G)}{p(S^{[0,T]}\mid\Lambda',G)}=\int \mathrm{d}p(S^{[0,T]}\mid\Lambda,G) \ln \frac{p(S^{[0,T]}\mid\Lambda,G)}{p(S^{[0,T]}\mid\Lambda',G)}.
\end{align*}
Inserting \eqref{eq:llh-ctbn} yields
\begin{align} \label{eq:KL_param}
&\KL{p(S^{[0,T]}\mid\Lambda,G)}{p(S^{[0,T]}\mid\Lambda',G)}=\sum_{n,x,x'\neq x,u} \\
&\left\{\Lambda'_{n}(x,x', u)-\Lambda_{n}(x,x', u)\right\}\expectation{}{\hat{T}_{n}(x,u)\mid \Lambda,G}
-\ln\frac{\Lambda_{n}(x,x', u)}{\Lambda'_{n}(x,x', u)}\expectation{}{\hat{M}_{n}(x,x',u)\mid \Lambda,G}\nonumber,
\end{align}
with the expectations being taken with respect to the process $p(S^{[0,T]}\mid\Lambda,G)$. The expected moments can not be calculated from the parametric form of $p(S^{[0,T]}\mid\Lambda,G)$ directly. Instead, we will construct an ODE for the moments of the CTMC recovered after amalgamation, and recover its expectation as solutions. The moments of the CTBN can then be calculated as projections of the CTMC moments, the dwelling times per state $T(s)$ and the number of transitions $M(s,s')$
\begin{align}\label{eq:projection-trans}
\hat{M}_{n}(x,x', u)&=\sum_{s,s'}M(s,s')\mathbb{1}(s_{n}'=x')\mathbb{1}(s_{n}=x)\mathbb{1}(s_{\mathrm{par}(n)}=u),\\ \label{eq:projection-dwell}
\hat{T}_{n}(x, u)&=\sum_{s}T(s)\mathbb{1}(s_{n}=x)\mathbb{1}(s_{\mathrm{par}(n)}=u).
\end{align}

\subsection{Moment ODEs of a CTMC}
\textbf{Expected Dwelling-times.}
The expected dwelling-times  $\expectation{}{T(s)}$ in a state $s\in\mathcal{S}$ of a CTMC are calculated as solution of an ODE. For this, we need to consider the evolution of the stochastic process $T(s,t)$, the dwelling times in state $s\in\mathcal{S}$ up to time $t$. For this process we can denote transition probabilities, by considering the dynamics of the CTMC
\begin{align*}
&p(T(s,t+h)=\tau+h\mid T(s,t)=\tau)=p(S(t+h)=s,S(t)=s),\\
&p(T(s,t+h)=\tau\mid T(s,t)=\tau)=1-p(S(t+h)=s,S(t)=s).
\end{align*}
Thus $T(s,t)$ evolves according to 
\begin{align*}
p(T(s,t+h)=\tau)&=p(S(t+h)=s,S(t)=s)p(T(s,t)=\tau-h)\\
&+\left[1-p(S(t+h)=s,S(t)=s)\right]p(T(s,t)=\tau).
\end{align*}
For small $h$, we can expand $p(T(s,t)=\tau-h)=p(T(s,t)=\tau)-h\partial_{\tau}p(T(s,t)=\tau)+o(h)$, and we arrive at 
\begin{align*}
\frac{p(T(s,t+h)=\tau)-p(T(s,t)=\tau)}{h}=-p(S(t)=s)\partial_{\tau}p(T(s,t)=\tau)+o(h).
\end{align*}
We can now take the expectation $\expectation{}{T(s,t)}=\int_0^\infty \mathrm{d} \tau'  \tau' p(T(s,t)=\tau')$, and the continuum limit $h\rightarrow 0$ in order to arrive at
\begin{align*}
\partial_{t}\mathrm{\mathsf{{E}}}\left[T(s,t)\right]=-p(S(t)=s)\int_{0}^{\infty}\mathrm{{d}}\tau'\,\tau'\partial_{\tau'}p(T(s,t)=\tau'),
\end{align*}
which, after integration by parts, reduces to simply
\begin{align*}
\partial_{t}\mathrm{\mathsf{{E}}}\left[T(s,t)\right]=p(S(t)=s).
\end{align*}
Thus, the expected dwelling-time is given by the solution
\begin{align}\label{eq:expected-dwell}
\mathrm{\mathsf{{E}}}\left[T(s)\right]=\int_{0}^{T}\mathrm{{d}}t\;p(S(t)=s).
\end{align}

\textbf{Expected Number of Transitions.}
Similarly to above, we can compute the expected number of transitions of a CTMC $\expectation{}{M(s,s')}$. The computation is analogous to above. We consider the stochastic process $M(s,s',t)$ of transitions from $s$ to $s'$ till time $t$. Transition probabilities are
\begin{align*}
&p(M(s,s',t+h)=k\mid M(s,s',t)=k-1)=p(S(t+h)=s',S(t)=s),\\
&p(M(s,s',t+h)=k\mid M(s,s',t)=k)=1-p(S(t+h)=s',S(t)=s).
\end{align*}
After inserting the identity $p(S(t+h)=s',S(t)=s)=\mathbb{1}(s=s')+hW(s,s')+o(h)$, we arrive at
\begin{align*}
&\frac{p(M(s,s',t+h)=k)-p(M(s,s',t)=k)}{h}\\
&=W(s,s')p(S(t)=s)\left[p(M(s,s',t)=k-1)-p(M(s,s',t)=k)\right]+o(h).
\end{align*}
The expected number of transitions can be calculated via $\expectation{}{M(s,s',t)}=\sum_{k=0}^\infty p(M(s,s',t)=k)$. Noticing that $p(M(s,s',t)=k-1)=0$ for $k<1$, we can perform an index-shift $k\rightarrow k+1$and arrive at
\begin{align*}
&\frac{\expectation{}{M(s,s',t+h)}-\expectation{}{M(s,s',t)}}{h}\\
&=W(s,s')p(S(t)=s)\left[\expectation{}{M(s,s',t)}-\expectation{}{M(s,s',t)}+1\right]+o(h),
\end{align*}
and thus in the continuum limit $h\rightarrow 0$ we recover the ODE,
\begin{align*}
\partial_t\expectation{}{M(s,s',t)}=W(s,s')p(S(t)=s),
\end{align*}
with the solution
\begin{align}\label{eq:expected-trans}
\expectation{}{M(s,s')}=W(s,s')\expectation{}{T(s)}.
\end{align}

\subsection{(V)BHC for Parameter Learning}
Equipped with the moments derived in the last Section, we can now derive the (V)BHC. The VBHC takes the form of an expected KL-divergence
\begin{align*}
\mathrm{{VBHC}}=&\int \mathrm{d}\Lambda\int \mathrm{d}\Lambda'\,p(\Lambda\mid\mathcal{H},G)q_\kappa(\Lambda')\KL{p(S^{[0,T]}\mid\Lambda,G)}{p(S^{[0,T]}\mid\Lambda',G)}\nonumber\\
&+\KL{q_\kappa(\Lambda)}{p(\Lambda\mid\mathcal{H},G)}\nonumber,
\end{align*}
with the KL given in appendix \eqref{eq:KL_param}. As explained in the main-text, we have ${p(\Lambda\mid\mathcal{H},G)=\prod_{n,x,x',u}\mathrm{Gam}\left(\Lambda_n(x,x', u)\mid \bar{\alpha}_n(x,x',u),\bar{\beta}_n(x, u)\right)}$ and choose ${q_\kappa(\Lambda')=\prod_{n,x,x',u}\mathrm{Gam}\left(\Lambda_n(x,x', u)\mid \alpha^{\kappa}_n(x,x',u),\beta^{\kappa}_n(x, u)\right)}$. As the expected moments in \eqref{eq:KL_param} only depend on $\Lambda$, we can calculate the integral over $\Lambda'$ analytically. For this, we notice that the moments 
\begin{align*}
    &\expectation{}{\Lambda_n(x,x',u)}=\frac{ \alpha^{\kappa}_n(x,x',u)}{\beta^{\kappa}_n(x, u)},\\
    &\expectation{}{\ln\Lambda_n(x,x',u)}=\psi^{(0)}({\alpha^{\kappa}_n(x,x',u)})-\ln{\beta^{\kappa}_n(x, u)},
\end{align*}
where the expectation is w.r.t $q_\kappa(\Lambda)$, have a closed form expression. By insertion into \eqref{eq:KL_param}, we recover the expression from the main-text. Finally, we notice that
\begin{align*}
&\KL{q_\kappa(\Lambda')}{p(\Lambda\mid\mathcal{H},G)}=\\
&\sum_{n,x,x',u}\mathrm{{KL}}\left(\mathrm{{Gam}}(\alpha^{\kappa}_n(x,x',u),\beta^{\kappa}_n(x,u))\,||\,\mathrm{{Gam}}(\bar{\alpha}_n(x,x',u),\bar{\beta}_n(x,u))\right)\nonumber,
\end{align*}
with the KL-divergence between two gamma-distributions~\cite{Soch2016}
\begin{align*}
&\mathrm{{KL}}\left(\mathrm{{Gam}}(\alpha^{\kappa}_n(x,x',u),\beta^{\kappa}_n(x,u))\,||\,\mathrm{{Gam}}(\bar{\alpha}_n(x,x',u),\bar{\beta}_n(x,u))\right)=\nonumber\\
&\bar{\alpha}_n(x,x',u)\ln\left(\frac{\beta^{\kappa}_n(x,u)}{\bar{\beta}_n(x,u)}\right)-\ln\left(\frac{\Gamma(\alpha^{\kappa}_n(x,x',u))}{\Gamma(\bar{\alpha}_n(x,x',u))}\right)\nonumber\\
&+(\alpha^{\kappa}_n(x,x',u)-\bar{\alpha}_n(x,x',u)))\psi(\alpha^{\kappa}_n(x,x',u)))-(\beta^{\kappa}_n(x,u)-\bar{\beta}_n(x,u))\frac{\alpha^{\kappa}_n(x,x',u)}{\beta^{\kappa}_n(x,u)}\nonumber.
\end{align*}
\textbf{Gradients.}
The gradients of the VBHC can be calculated in (semi-)analytical form
\begin{align}\label{eq:gradients-alpha}
\partial_{\alpha^{\kappa}_n(x,x',u)}\mathrm{{VBHC}}&=\int d\Lambda\:p(\Lambda\mid\mathcal{H},G)\expectation{}{\hat{T}_n\mid \Lambda,G}\left\{ \Lambda\psi^{(1)}(\alpha^{\kappa}_n(x,x',u))-\frac{1}{\beta^{\kappa}_n(x,u)}\right\} \\
&+\alpha^{\kappa}_n(x,x',u)\psi^{(1)}(\alpha^{\kappa}_n(x,x',u))-\frac{(\beta^{\kappa}_n(x,u)-\bar{\beta}_n(x,u))}{\beta^{\kappa}_n(x,u)}\nonumber\\
\partial_{\beta^{\kappa}_n(x,u)}\mathrm{{VBHC}}&=\int d\Lambda\:p(\Lambda\mid\mathcal{H},G)\expectation{}{\hat{T}_n\mid \Lambda,G}\left\{ \frac{\alpha^{\kappa}_n(x,x',u)}{\beta^{\kappa}_n(x,u)^{2}}-\frac{\Lambda}{\beta^{\kappa}_n(x,u)}\right\}\label{eq:gradients-beta}\\ &+\frac{\bar{\alpha}_n(x,x',u)}{\beta^{\kappa}_n(x,u)}-\frac{\alpha^{\kappa}_n(x,x',u)}{\beta^{\kappa}_n(x,u)}+(\beta^{\kappa}_n(x,u)-\bar{\beta}_n(x,u))\frac{\alpha^{\kappa}_n(x,x',u)}{\beta^{\kappa}_n(x,u)^{2}}\nonumber.
\end{align}
If necessary, also higher-order derivatives can be computed in principle.

In all results above, the corresponding BHC expressions are recovered by setting $\beta^{\kappa}_n(x,u)=\bar{\beta}_n(x,u)$ and $\alpha^{\kappa}_n(x,x',u)=\bar{\alpha}_n(x,x',u)$.

\subsection{(V)BHC for Structure Learning}

\textbf{KL-divergence between Marginal CTBNs.}
The marginal likelihood of a path $\hat{S}^{[0,T]}\sim p(\hat{S}^{[0,T]}\mid \Lambda,G)$, with statistics $\hat{T}_{n}(x, u)$ and $\hat{M}_{n}(x,x',u)$, given a structure and history $\mathcal{H}$ can be calculated via marginalization of \eqref{eq:llh-ctbn}
\begin{align}\label{eq:marg-llh-ctbn}
p(\hat{S}^{[0,T]}\mid G,\mathcal{H})&=\int d\Lambda\:p(\Lambda\mid \mathcal{H},G)\prod_{n}\prod_{x,x',u}\Lambda_{n}(x,x', u)^{\hat{M}_{n}(x,x',u)}\exp{\left\{ -\Lambda_{n}(x,x',u)+\hat{T}_{n}(x, u)\right\} }\nonumber\\
&\propto\prod_n\prod_{x,x'\neq x,u}\Gamma(\hat{\alpha}_n(x,x',u))\hat{\beta}_n(x,u)^{-\hat{\alpha}_n(x,x',u)},
\end{align}
where $\hat{\alpha}_n(x,x',u)=\hat{M}_{n}(x,x', u)+\bar{\alpha}_n(x,x',u)$ and $\hat{\beta}_n(x,u)=\hat{T}_{n}(x, u)+\bar{\beta}_n(x,u)$ and ${\bar{\alpha}_n(x,x',u)=\alpha_n(x,x',u)+M_n(x,x',u,i=0)}$ and ${\bar{\beta}_n(x,x',u)=\beta_n(x,u)+T_n(x,u,i=0)}$, see main-text.
The KL between two measures, in this case CTBNs with different graphs, is defined via the integration over all paths
\begin{align*}
\KL{p(S^{[0,T]}\mid G,\mathcal{H})}{p(S^{[0,T]}\mid G',\mathcal{H})}=\int \mathrm{d}p(S^{[0,T]}\mid G,\mathcal{H}) \ln \frac{p(S^{[0,T]}\mid G,\mathcal{H})}{p(S^{[0,T]}\mid G',\mathcal{H})}\nonumber.
\end{align*}
In order to avoid solving the computationally taxing solution of the marginal master-equation~\cite{Studer2016,Linzner2018} (which is an integro-differential equation),
we can express this in terms of the original path-measure
\begin{align*}
&\KL{p(S^{[0,T]}\mid G,\mathcal{H})}{p(S^{[0,T]}\mid G',\mathcal{H})}=\\
&\int\mathrm{d}p(\Lambda\mid G,\mathcal{H}) \int \mathrm{d}p(S^{[0,T]}\mid \Lambda, G) \ln \frac{p(S^{[0,T]}\mid G,\mathcal{H})}{p(S^{[0,T]}\mid G',\mathcal{H})}\nonumber.
\end{align*}

Inserting \eqref{eq:marg-llh-ctbn} yields
\begin{align} \label{eq:KL_struct}
&\KL{p(S^{[0,T]}\mid G,\mathcal{H})}{p(S^{[0,T]}\mid G',\mathcal{H})}=\int\mathrm{d}p(\Lambda\mid G,\mathcal{H}) \int \mathrm{d}p(S^{[0,T]}\mid\Lambda, G)\sum_{n\in \aleph}\\
&\sum_{u\in \mathcal{U}_n^G}\sum_{u'\in \mathcal{U}_n^{G'}}\sum_{x,x'\neq x} \left[\ln\frac{\Gamma(\hat{\alpha}_{n}(x,x', u\,))}{\Gamma(\hat{\alpha}_{n}(x,x', u'))}
+\hat{\alpha}_{n}(x,x',u')\ln\hat{\beta}_{n}(x, u')-\hat{\alpha}_{n}(x,x',u)\ln\hat{\beta}_{n}(x, u)\right].\nonumber
\end{align}
As mentioned in the main-text, exact computation of the integral w.r.t $p(S^{[0,T]}\mid\Lambda, G)$ is not feasible, due to non-linearity. For this reason we expand this KL around the expected transitions and dwelling times and arrive at
\begin{align} \label{eq:KL_struct_approx}
&\KL{p(S^{[0,T]}\mid G,\mathcal{H})}{p(S^{[0,T]}\mid G',\mathcal{H})}\approx\int\mathrm{d}p(\Lambda\mid G,\mathcal{H})\sum_{n\in \aleph}\sum_{u\in \mathcal{U}_n^G}\sum_{u'\in \mathcal{U}_n^{G'}}\sum_{x,x'\neq x} \\
& \left[\ln\frac{\Gamma(\mathsf{E}[\hat{\alpha}_{n}(x,x',u\,)])}{\Gamma(\mathsf{E}[{\hat{\alpha}}_{n}(x,x',u')])}
+\mathsf{E}[{\hat{\alpha}}_{n}(x,x',u')]\ln\mathsf{E}[\hat{\beta}_{n}(x, u')]-\mathsf{E}[\hat{\alpha}_{n}(x,x',u\,)\ln\mathsf{E}[\hat{\beta}_{n}(x, u)]\right],\nonumber\\
&\equiv \mathcal{F}[\kappa,p(G\mid \mathcal{H})] \nonumber
\end{align}
with ${\mathsf{E}[\hat{\alpha}_{n}(x,x',u)]\equiv{\bar{\alpha}}_{n}(x,x',u)+\mathsf{E}[\hat{M}_n(x,x',u)\mid \Lambda,G]}$ and ${\mathsf{E}[{\hat{\beta}}_{n}(x,u)]\equiv{\bar{\beta}}_{n}(x,u)+\mathsf{E}[\hat{T}_n(x,u)\mid \Lambda,G]}$.

Below, we derive higher-order moments of the transitions and dwelling-times. This allows to compute higher-order approximations of this KL-divergence, under higher computational costs. However, in this work a first order approximation was sufficient to demonstrate effectiveness of our method.

\textbf{VBHC for Structure Learning.}
We can then approximate the VBHC by 
\begin{align*} 
\mathrm{VBHC}\approx \mathcal{F}[\kappa,p(G\mid \mathcal{H})]+\KL{q_\kappa(G)}{p(G\mid \mathcal{H})}, \end{align*}
with the KL-divergence, between two categoricals
\begin{align*} 
\KL{p(G\mid \mathcal{H})}{q_\kappa(G)}=\sum_{G}{q_\kappa(G)}\left(\ln{q_\kappa(G)}-\ln p(G\mid \mathcal{H})\right).
\end{align*}
While the form of $\mathcal{F}$ is compact, it is  helpful for computational reasons to re-order this summation into a node-wise form. This is helpful, as it will allow is to compute sample approximations of the VBHC, where only a summation over local parent-sets instead of global graphs needs to be performed
\begin{align*} 
&\mathcal{F}[\kappa,p(G\mid \mathcal{H})]= \sum_{n\in \aleph}\sum_{G,G'}p(G\mid \mathcal{H})q(G')\int\mathrm{d}p(\Lambda\mid G,\mathcal{H})\sum_{u\in \mathcal{U}_n^G}\sum_{u'\in \mathcal{U}_n^{G'}}\sum_{x,x'\neq x} \\
& \left[\ln\frac{\Gamma(\mathsf{E}[\hat{\alpha}_{n}(x,x',u\,)])}{\Gamma(\mathsf{E}[{\hat{\alpha}}_{n}(x,x',u')])}
+\mathsf{E}[{\hat{\alpha}}_{n}(x,x',u')]\ln\mathsf{E}[\hat{\beta}_{n}(x, u')]-\mathsf{E}[\hat{\alpha}_{n}(x,x',u\,)\ln\mathsf{E}[\hat{\beta}_{n}(x, u)]\right].\nonumber
\end{align*}
The product form of $\eqref{eq:marg-llh-ctbn}$ translates to a product posterior, if not broken by the prior, over parent-sets
\begin{align}\label{eq:par-post}
    p(G\mid \mathcal{H}) &= \prod_n p(\mathrm{par}^G(n)\mid\mathcal{H})\\
    &\,\,\propto  \prod_n p(\mathrm{par}^G(n))\prod_{x,x'\neq x}\prod_{u\in\mathcal{U}_n^{G}}\Gamma(\bar{\alpha}_n(x,x',u))\bar{\beta}_n(x,u)^{-\bar{\alpha}_n(x,x',u)}\nonumber.
\end{align}
This allows us to rewrite 
\begin{align*}
&\sum_{n\in \aleph}\sum_{G,G'} p(G)q_\kappa(G')=\\
&\sum_{n\in \aleph} \sum_{\mathrm{par}(n),\mathrm{par}'(n)\subset \mathcal{V}}p(\mathrm{par}(n) \mid \mathcal{H})q_\kappa(\mathrm{par}'(n))\sum_{G\mid \mathrm{par}(n),G'\mid \mathrm{par}'(n)}p(G\mid \mathcal{H})q_\kappa(G').
\end{align*}
We then get the form of the VBHC for structure learning, as used in algorithm 2
\begin{align}\label{eq:VBHC-struct-sample} 
&\mathcal{F}[\kappa,p(G\mid \mathcal{H})]= \sum_{n\in \aleph}\sum_{\mathrm{par}(n),\mathrm{par}'(n)}p(\mathrm{par}(n)\mid \mathcal{H})q_\kappa(\mathrm{par}'(n))\\
&\sum_{u\in \mathcal{U}_n^{\mathrm{par}(n)}}
\sum_{u'\in \mathcal{U}_n^{\mathrm{par}(n)'}}\sum_{G\mid \mathrm{par}(n)}
\sum_{G'\mid \mathrm{par}'(n)}
p(G\mid \mathcal{H})q_\kappa(G')\int\mathrm{d}p(\Lambda\mid G,\mathcal{H})
\sum_{x,x'\neq x}\nonumber \\
& \left[\ln\frac{\Gamma(\mathsf{E}[\hat{\alpha}_{n}(x,x',u\,)])}{\Gamma(\mathsf{E}[{\hat{\alpha}}_{n}(x,x',u')])}
+\mathsf{E}[{\hat{\alpha}}_{n}(x,x',u')]\ln\mathsf{E}[\hat{\beta}_{n}(x, u')]-\mathsf{E}[\hat{\alpha}_{n}(x,x',u\,)\ln\mathsf{E}[\hat{\beta}_{n}(x, u)]\right].\nonumber
\end{align}
Similarly, we make the ansatz for ${q_\kappa(G)= \prod_n q_\kappa(\mathrm{par}^G(n))}$, then the KL-divergence decomposes
\begin{align*} 
\KL{p(G\mid \mathcal{H})}{q_\kappa(G)}=\sum_n\sum_{\mathrm{par}(n)}{q_\kappa(\mathrm{par}(n))}\left(\ln{q_\kappa(\mathrm{par}(n))}-\ln p(\mathrm{par}(n)\mid \mathcal{H})\right).
\end{align*}
\textbf{Gradients.}
The gradient for the parameter $q_{\kappa}(\mathrm{par}'(n))$ can be calculated to be 
\begin{align}\label{eq:grad-VBHC-struct-sample} 
&\partial_{q_{\kappa}(\mathrm{par}'(n))}\mathrm{VBHC}=1+\ln{q_\kappa(\mathrm{par}'(n))}-\ln p(\mathrm{par}'(n)\mid \mathcal{H})+ \sum_{\mathrm{par}(n)}p(\mathrm{par}(n)\mid \mathcal{H})\\
&\sum_{u\in \mathcal{U}_n^{\mathrm{par}(n)}}
\sum_{u'\in \mathcal{U}_n^{\mathrm{par}(n)'}}\sum_{G\mid \mathrm{par}(n)}
\sum_{G'\mid \mathrm{par}'(n)}
p(G\mid \mathcal{H})q_\kappa(G')\int\mathrm{d}p(\Lambda\mid G,\mathcal{H})
\sum_{x,x'\neq x}\nonumber \\
& \left[\ln\frac{\Gamma(\mathsf{E}[\hat{\alpha}_{n}(x,x',u\,)])}{\Gamma(\mathsf{E}[{\hat{\alpha}}_{n}(x,x',u')])}
+\mathsf{E}[{\hat{\alpha}}_{n}(x,x',u')]\ln\mathsf{E}[\hat{\beta}_{n}(x, u')]-\mathsf{E}[\hat{\alpha}_{n}(x,x',u\,)\ln\mathsf{E}[\hat{\beta}_{n}(x, u)]\right].\nonumber
\end{align}

%\subsection{Relation of the VBHC to an MI upper-bound}
%We show the VBHC can be related to the MI upper-bound presented, among others, in~\cite{Poole2019,Foster2019}
%\begin{align*}
%\mathsf{{EIG}}=\int d {D}\int d\Theta\,p(\Theta)p( {D}\mid\Theta)\ln\left(\frac{p( {D}\mid \Theta)}{p( {D})}\right)=\int d {D}\int d\Theta\,p( {D},\Theta)\ln\left(\frac{p( {D},\Theta)}{p( {D})p(\Theta)}\right)=\mathrm{{MI}}
%\end{align*}
%now use the variational lower-bound
%\begin{align*}
%&\ln p(D)=\mathrm{{KL}}\left[q_{\kappa}(\Theta)||p(\Theta\mid  {D})\right]-\mathrm{{KL}}\left[q_{\kappa}(\Theta)||p(\Theta)\right]+\mathsf{{E}}\left[\ln p( {D}\mid\Theta)\right]
%\end{align*}
%inserting this into the mutual information, and bounding $\mathrm{{KL}}\left[q_{\kappa}(\Theta)||p(\Theta\mid  {D})\right]\geq 0$, yields an upper-bound
%\begin{align*}
%&\mathrm{{MI}}\leq\int d {D}\int d\Theta\,p( {D},\Theta)\left[\ln\left(p( {D},\Theta)\right)-\ln\left(p(\Theta)\right)+\mathrm{{KL}}\left[q_{\kappa}(\Theta)||p(\Theta)\right]-\mathsf{{E}}\left[\ln p( {D}\mid\Theta)\right]\right]
%\end{align*}
%with the expectations w.r.t $q_\kappa(\Theta)$. As $\exp\left\{\mathsf{{E}}\left[\ln p( {D}\mid\Theta)\right]\right\} \leq \mathsf{{E}}\left[\exp\left\{\ln p( {D}\mid\Theta)\right]\right\}$, we can further upper-bound 
%\begin{align*}
%\mathrm{{MI}}\leq\int d {D}\int d\Theta\,p( {D},\Theta)\left[\ln\left(\frac{p( {D},\Theta)}{p( {\Theta})q_{\kappa}( {D})}\right)\right]+\mathrm{{KL}}\left[q_{\kappa}(\Theta)||p(\Theta)\right],
%\end{align*}
%where we defined
%$
%q_{\kappa}( {D})\equiv \mathsf{{E}}\left[ p( {D}\mid\Theta)\right].
%$
\begin{figure}[t]
  \begin{center}
    \includegraphics[width=0.6\textwidth]{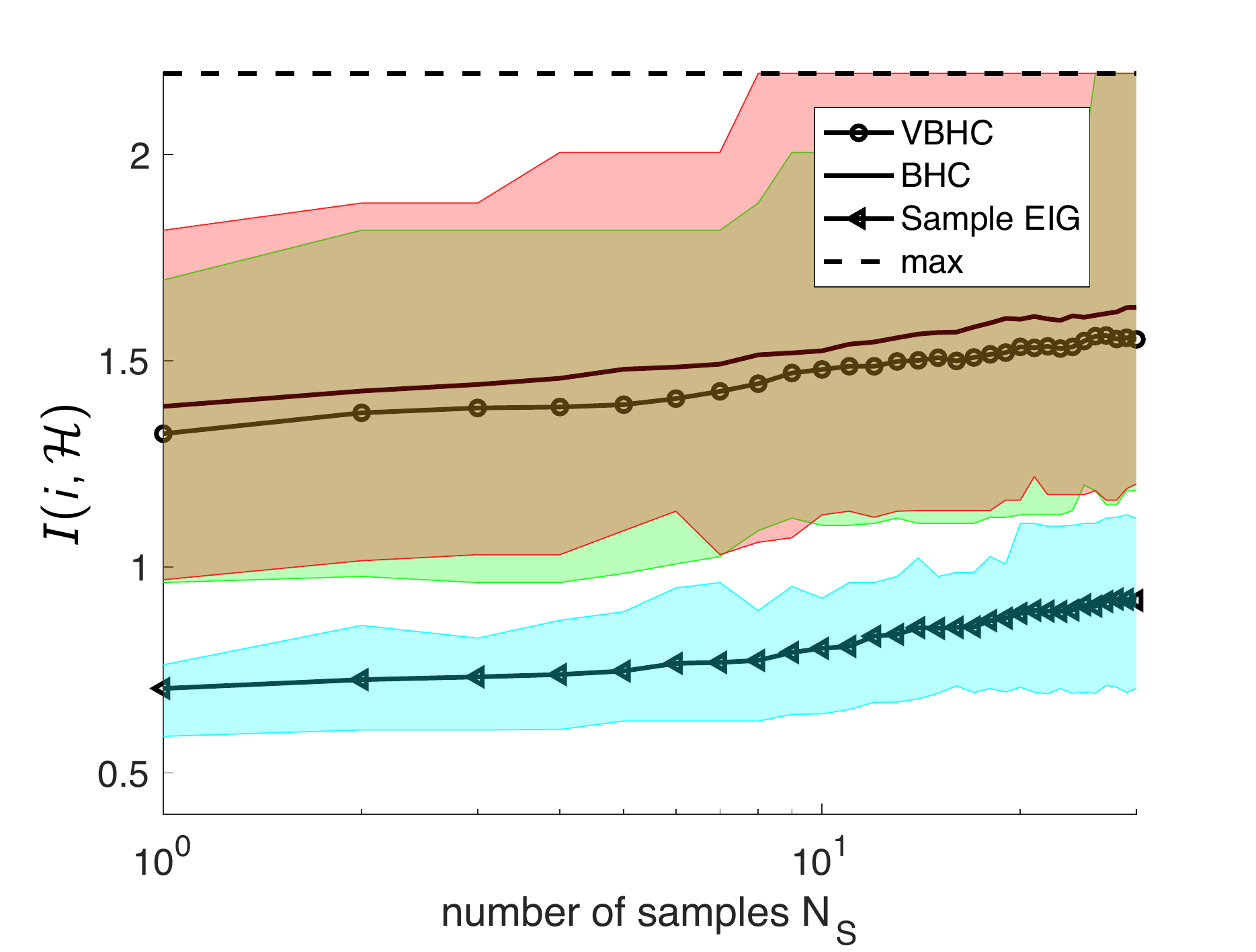}
  \end{center}
  \caption{Mutual information between design sample estimates and recommended interventions for different number of samples $N_S$. Areas denote 25-75$\%$ confidence intervals. }
   \label{fig:MI_design}
\end{figure}
\section{Experiments}
\subsection{Additional Experiments}
\textbf{Sample Estimates of Design Criteria.}
We want to investigate the viability of using sample estimates of different criteria for active learning of CTBNs. One basic requirement on such an estimate is that its recommendations actually depend on the history of observations  $\mathcal{H}\underset{\mathrm{design}}{\longrightarrow}i$. We can make this formal by the following non-parametric dependency check: The recommended intervention $i$ is dependent on experimental sequence $\mathcal{H}$ if they share high mutual information $I(i,\mathcal{H})$. We stress, that this does not reflect the quality of recommended interventions! We calculate the MI for random graphs of size $L=3$ for different sample sizes $N_S$ for random histories $\mathcal{H}$ consisting of 30 trajectories drawn from our synthetic network. The results are displayed in figure \ref{fig:MI_design}. For all sample sizes considered, (V)BHC shares a much higher MI with their recommended interventions, than the sample estimate of the EIG.
\begin{figure}[t]
  \begin{center}
    \includegraphics[width=0.9\textwidth]{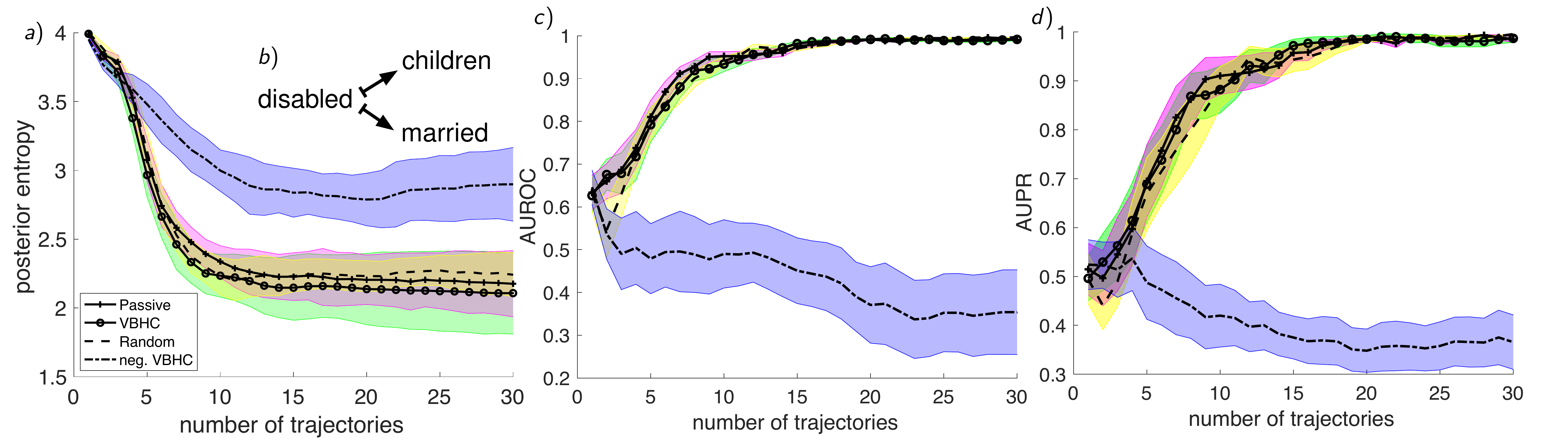}
  \end{center}
  \caption{a) Mean and variance (area) of the evolution of the posterior entropy in BHPS data-set for 100 repetitions. b) Sketch of the underlying network. c) AUROC and d) AUPR converge equally fast to the inferred network b) for all criteria but negative VBHC. }
   \label{fig:bhps_root}
\end{figure}
\subsection{Processing of British-Household Data-set}
As mentioned in the main-text, the British-Household Data-set is incomplete, as no complete paths of variables are provided, but only their measurement at singular time-points $t_i\in\{1,\dots,11\}$ (yearly for 11 years). In order to process, this kind of data, we employ a standard forward backward filter for continuous-time Markov jump processes, as in~\cite{Opper2001,Opper2008,Cohn2010,Linzner2018}. For this data $Y^{[0,T]}\equiv\{Y(t_i)\mid t_i\in\{1,\dots,11\}\}$ and $Y(t_i)\sim p(Y(t_i)\mid S(t_i))$ some observation model, with measurements at singular time-points, posterior inference of the marginals $p(S(t)=s\mid Y^{[0,T]})$ is implemented by solving a time-dependent master-equation
\begin{align*}%\label{eq:post-m-eq}
\nonumber &\frac{\mathrm d}{\mathrm d t}p(S(t)=s\mid Y^{[0,T]}) = \\
\nonumber  &\sum_{s'\neq s}\left[\hat{W}(s',s,t)p(S(t)=s\mid Y^{[0,T]})-\hat{W}(s,s',t)p(S(t)=s'\mid Y^{[0,T]})\right]
\end{align*}
with $\hat{W}(s,s',t)=W(s,s')\frac{\rho(s',t)}{\rho(s,t)}$ and
\begin{align*}
 \frac{\mathrm d}{\mathrm d t}\rho(s',t) &= -\sum_{s'\neq s}\left[W(s',s)\rho(s,t)-W(s,s')\rho(s',t)\right]\\
 \text{subject to: } \lim_{t\rightarrow t^{-}_i}\rho(s,t)&=\lim_{t\rightarrow t^{+}_i}\rho(s,t)\ln p(Y(t_i)\mid S(t_i)=s).
\end{align*}
This allows to calculate the marginal likelihood 
\begin{align*}
p(Y^{[0,T]}\mid W)=\prod_{s,s'\neq s}W(s,s')^{\expectation{}{M(s,s')\mid Y^{[0,T]}}}\exp\left\{W(s,s)\expectation{}{T(s)\mid Y^{[0,T]}}\right\},
\end{align*}
with $\expectation{}{T(s)\mid Y^{[0,T]}}\equiv \int \mathrm{d}t\;p(S(t)=s\mid Y^{[0,T]})$ and $\expectation{}{M(s,s')\mid Y^{[0,T]}}\equiv W(s,s')\expectation{}{T(s)\mid Y^{[0,T]}}$.
By calculation of the corresponding moments of the CTBNs by appendix ~\eqref{eq:projection-dwell} and~\eqref{eq:projection-trans}, we can also write this likelihood in terms of rates $\Lambda$ and structure $G$ 
\begin{align*}
&p(Y^{[0,T]}\mid \Lambda,G)=\\
&\prod_{n,x,x'\neq x,u}\Lambda_n(x,x',u)^{\expectation{}{M_n(x,x',u)\mid Y^{[0,T]}}}\exp\left\{\Lambda_n(x,x,u)\expectation{}{T_n(x,u)\mid Y^{[0,T]}}\right\}.
\end{align*}
As can be seen in~\cite{Linzner2018}, this finally allows to form a posterior over parameters ${p(\Lambda\mid Y^{[0,T]})\propto p(Y^{[0,T]}\mid \Lambda)p(\Lambda)}$, which is again a Gamma distribution, if $p(\Lambda)$ is gamma-distributed. Similarly, this holds for structures, by marginalization. Aside from this posterior calculation, everything about our method remains the same for incomplete data.

In \cref{fig:bhps_root} a), we track the evolution of the posterior entropy over structures for 100 independent runs. In \cref{fig:bhps_root} b) and c), we show that for all designs (except the "worst" design neg. VBHC) the inferred network converges against the one inferred using the full data-set (using AUROC and AUPR as metrics). We note that the effect of active learning can be expected to be small in this synthetic scenario, as we were only able to intervene on a single node.

\end{document}